\documentclass[journal]{IEEEtran}

 \usepackage{graphics,subfigure}

\usepackage{epsfig,xcolor}
\usepackage{array}
\usepackage{mdwmath}
\usepackage{mdwtab}
\usepackage{eqparbox}
\usepackage{multirow}
\usepackage[cmex10]{amsmath}
\usepackage{url}
\usepackage{graphicx}
\usepackage{subfigure}
\usepackage[absolute,showboxes]{textpos}
\usepackage{amssymb}

\begin{document}

\title{A fast and accurate iris segmentation method using an LoG filter and its zero-crossings}
\author{Tariq M. ~Khan,
        Donald G. ~Bailey,
        and Yinan ~Kong
\thanks{Tariq M. Khan and Yinan Kong are with Department of Engineering, Macquarie University, Sydney, Australia }
\thanks{Donald G. Bailey is with the School of Engineering and Advanced Technology, Massey University, Palmerston North, New Zealand }}
\maketitle
\begin{abstract}
This paper presents a hybrid approach to achieve iris localization based on a Laplacian of Gaussian (LoG) filter, region growing, and zero-crossings of the LoG filter. In the proposed method, an LoG filter with region growing is used to detect the pupil region. Subsequently, zero-crossings of the LoG filter are used to accurately mark the inner and outer circular boundaries. The use of LoG based blob detection along with zero-crossings makes the inner and outer circle detection fast and robust. The proposed method has been tested on three public databases: MMU version 1.0, CASIA-IrisV1 and  CASIA-IrisV3- Lamp. The experimental results demonstrate the segmentation accuracy of the proposed method. The robustness of the proposed method is also validated in the presence of noise, such as eyelashes, a reflection of the pupil, Poisson, Gaussian, speckle and salt-and-pepper noise. The comparison with well-known methods demonstrates the superior performance of the proposed method's accuracy and speed.
\end{abstract}
\IEEEpeerreviewmaketitle

\section{Introduction}
\label{sec:intro}  
\begin{figure*}
  \centering
  \includegraphics[scale=1.3]{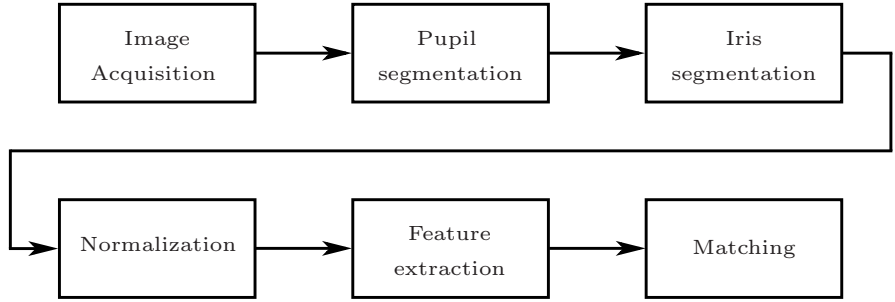}\\
  \caption{Block diagram of an iris recognition system}\label{IrisBlock}
\end{figure*}

The use of fraudulent identities is considered to be a key enabler of serious organised crime and even terrorism \cite{khan2020real, abbas2019fast, sabir2020reducing}. Biometrics is a fast developing science that can provide a higher level of security, convenience, and efficiency to protect against identity theft than traditional password-based methods for user authentication \cite{khan2016spatial, khan2016efficient, khan2016fusion}. Humans have many biometric traits such as a face, hand geometry, fingerprint, voice and iris \cite{khan2011automatic}, that can be used for identity verification. However, iris recognition is found to be accurate and one of the more reliable methods due to its high degree of uniqueness and randomness, even between identical twins, and remains constantly stable throughout an adult's life \cite{ibrahim2012iris, ibrahim2011novel}.\\
\indent The iris is a nearly circular shaped region between the sclera and the pupil. It is an internal organ that is well protected from the environment and can be seen from outside the body \cite{Ross2012}. It consists of many features like furrows, freckles, stripes, coronas, ligaments, arching, zigzag collarette, ridges, rings, and crypts \cite{Daugm1993, Wildes1997, Ma2004}. These features are statistically stable, unique, and are randomly distributed in the human iris region \cite{Daugm2007, Liam2002}. These properties make the iris a secure and reliable source of personal identification \cite{khan2015hardware}.\\
\indent Generally, the essential steps within an iris recognition system are: eye image acquisition, segmentation of the inner and outer boundary of iris, extraction of unique features, feature matching and finally the recognition of a person \cite{ibrahim2012iris}, as shown in Fig. \ref{IrisBlock}. Of these steps, the segmentation of the inner and outer boundary of iris plays a vital role towards system accuracy. Iris segmentation localises two different boundaries. First, it segments out the pupil's outer boundary, known as the pupillary or inner boundary of the iris, and then the outer or limbic boundary of the iris. Iris segmentation is a computationally intensive task in iris recognition \cite{ibrahim2012iris}.\\
\indent Although state-of-the-art methods \cite{Daugm2004, Boles1998, Sun2004, Yu2007} are very effective for iris recognition, their performance is greatly affected by iris segmentation. The reasons are as follows:
\begin{itemize}
 \item Iris segmentation defines the contents of the features that are subsequently processed, by normalisation, feature extraction, and matching. Thus the accuracy of iris recognition is directly related to iris segmentation. Inaccurate iris segmentation is reported to be the main cause of failure in iris recognition systems \cite{Ma2004}.
  \item Processing speed is a bottleneck in practical applications, while in iris recognition the most time-consuming module is iris segmentation \cite{Daugm2004, He2006}.
\end{itemize}
\indent There are several challenges in practical iris segmentation. Some of them are the image acquisition angle, pupil dilation, occlusion, focus, and image clarity. Pupillary and limbic boundaries are nonmuscular, which can lead to inaccuracy if fitted with simple shape assumptions. All of these challenges makes the iris segmentation process difficult. Therefore a fast, accurate and robust iris segmentation algorithm is highly desirable.\\
\indent In the literature, different techniques are proposed for fast and accurate iris segmentation. These can be divided into two broad categories: Shape-based detectors \cite{Wildes1997, Daugm2004} and intensity-based thresholding \cite{khan2011automatic, ibrahim2012iris, ibrahim2011novel}. Generally, shape-based detection gives better accuracy but is slower in processing \cite{khawaja2019multi, naveed2021towards, khan2022width}. On the other hand, intensity-based threshold methods are fast, but they are less accurate than the shape-based detectors. \\
\indent In this chapter, we investigate the combination of a fast shape-based detector with an intensity-based threshold to accurately segment an iris. A Laplacian of Gaussian (LoG) filter is used as a shape detector along with region growing and an intensity-based threshold is efficiently used to locate the true pupil region. Then the zero-crossings of the LoG filter are used to mark the true inner and outer boundaries of the iris. The combination of these filters not only gives better accuracy than intensity-based threshold methods but also gives much better processing speed than shape-based detectors.
\indent The rest of this paper is organized as follows.Background and related work is presented in Section \ref{ch3:sec1}. Section \ref{ch3:sec2} give details of proposed method. Experimental results are presented in Section \ref{ch3:sec3}. Computational cost and limitation are discussed in section \ref{ch3:sec4} followed by the conclusion of this chapter in Section \ref{ch3:sec5}.
\section{Background and related work}
\label{ch3:sec1}
In iris segmentation, the first step is to find the pupil centre and then to fit a circle to localise the pupil boundaries \cite{ibrahim2012iris}. The centre of the iris lies within the pupil, but the two circles are not concentric. An accurate localization of the pupil reduces the search space for the centre of the iris. Duagman proposed an edge-based detector using an integral differential operator (IDO) \cite{Daugm2002}. The operator searches over the image domain $ (x,y)$ for the maximum in the blurred partial derivatives with respect to increasing radius $r$, of the normalised contour integral of $I(x,y)$. The complete operator behaves as a circular edge detector, blurred at a scale set by $\sigma$, searching iteratively for the maximal contour integral. A circular Hough transform is used to detect the inner and outer boundaries of the iris. Three parameters $(x_0,y_0,r)$ are used to define each circle, where $(x_0,y_0)$ is the center and $r$ is the radius of the circle. Similarly, a mixture of the gradient edge-map and the circular Hough transform was used by Wildes \cite{Wildes1997} to pinpoint the iris boundaries.\\
\indent The literature provides evidence that histogram and threshold based techniques are faster than using the Hough transform \cite{Bowyer2008}. In histogram-based techniques, thresholding is used for locating the pupil considering it as the darkest region in an eye image \cite{khan2011automatic}. For pupil detection, Zhang \cite{XU2007} first divided the eye image into small rectangular blocks of fixed size and then found the average intensity value of each block. Dey \cite{Dey2008} used down-sampling on the input image before pupil and iris localization. To find all the connected components, first, contrast scratching is applied to the  down-sampled image followed by the image binarization.\\
\indent Ibrahim \emph{et al.} \cite{ibrahim2012iris} used histogram-based and standard deviation based adaptive thresholding to localise the pupillary boundary. A range of grey levels that has the highest probability of the pupil is found by moving a circular window. The window that contains a grey level peak with a minimum standard deviation of x- and y- coordinates is selected as the pupil region. This technique may fail for images containing other objects such as eyelashes, eyebrows, hair, and possibly the black-frame of glasses. Similarly, Khan \emph{et al.} \cite{khan2011automatic} used histogram-based thresholding along with eccentricity to extract the pupillary boundary. Their technique lacks the ability to handle multiple objects in a given binary image. Use of eccentricity on its own could be misleading if a small object (other than the pupil) has smaller eccentricity in an image. To overcome these issues, Jan \emph{et al.} \cite{Jan2013} proposed a technique in which a common eye position is generalised by using integral image projection functions. Then it was binarized by using an adaptive threshold via a histogram bisection method. It is controlled by a parameter vector that is recorded by eccentricity, image statistics, and the two-dimensional (2D) object geometry. Then the limbic boundary is identified in the horizontal direction by using gradient information. Lastly, the iris boundaries are localised by using a mixture of radial gradients. This method gives effective results for iris databases but with certain limitations. First, the dark locations (e.g. pupil, etc.) were highlighted because of the property of integral image projection. If other low-intensity regions (e.g., eyebrows, eyelashes, hair, and the frame of glasses) block the dark regions (e.g. pupil, etc.), then the integral image projection may fail to highlight it because of the coordination failure of the horizontal and vertical projection functions. Secondly, the useful combination of area and eccentricity of 2D objects are used to locate the pupil in a binary image but, as was discussed earlier, the results are better for a perfectly round pupil, but method may not provide the desired results for a distorted pupil, for example, CASIA- IrisV3-Interval and CASIA-IrisV4-Thousand iris databases.\\
\indent Although in literature, many different iris segmentation methods have been proposed, many of them are either computationally or present relatively high or unacceptable segmentation error rates. The iris image segmentation algorithm proposed in this chapter consists of two major modules, namely pupil detection, and limbic boundary localization including eyelid/eyelash detection. The implemented algorithms avoid unnecessary processing over image regions that do not contain relevant information for iris image segmentation, and consequently iris recognition.\\

\section{Proposed method}
\label{ch3:sec2}
We have proposed a two-stage method for iris localization. In stage 1, the pupil (iris inner boundary) is localised, an important step in iris segmentation for two reasons:
\begin{enumerate}
  \item In the iris images, if the pupil is wrongly localised, then there are often errors in detecting the limbic boundary, as the iris's outer boundary localization methods use the pupil circle parameters as inputs \cite{Jan2013,Wang2014, A.Radman2013,khan2011automatic}.
  \item The time consumed in pupil localization is much more than for the limbic boundary localization because the whole eye (iris) image is processed in pupil localization,  whereas a sub-image can be processed for the limbic boundary localization \cite{Wang2014, A.Radman2013}.
\end{enumerate}
In stage 2, the limbic boundary is localised. The details of the proposed method are given in the next subsections.
\subsection{Pupillary boundary localization}
\begin{figure*}[h]
 	\begin{center}
 		\subfigure[] {\label{compasiron:-a}\includegraphics[scale=0.3]{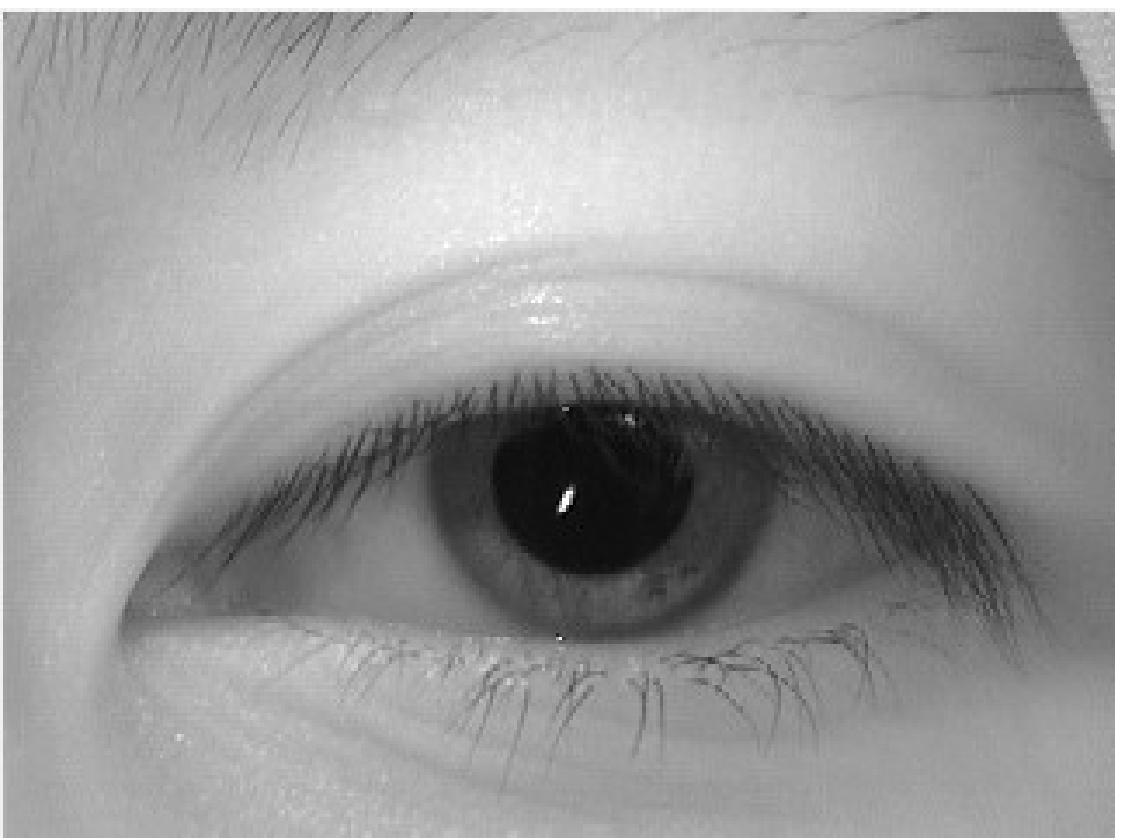}}
 		\subfigure[] {\label{compasiron:-b}\includegraphics[scale=0.3]{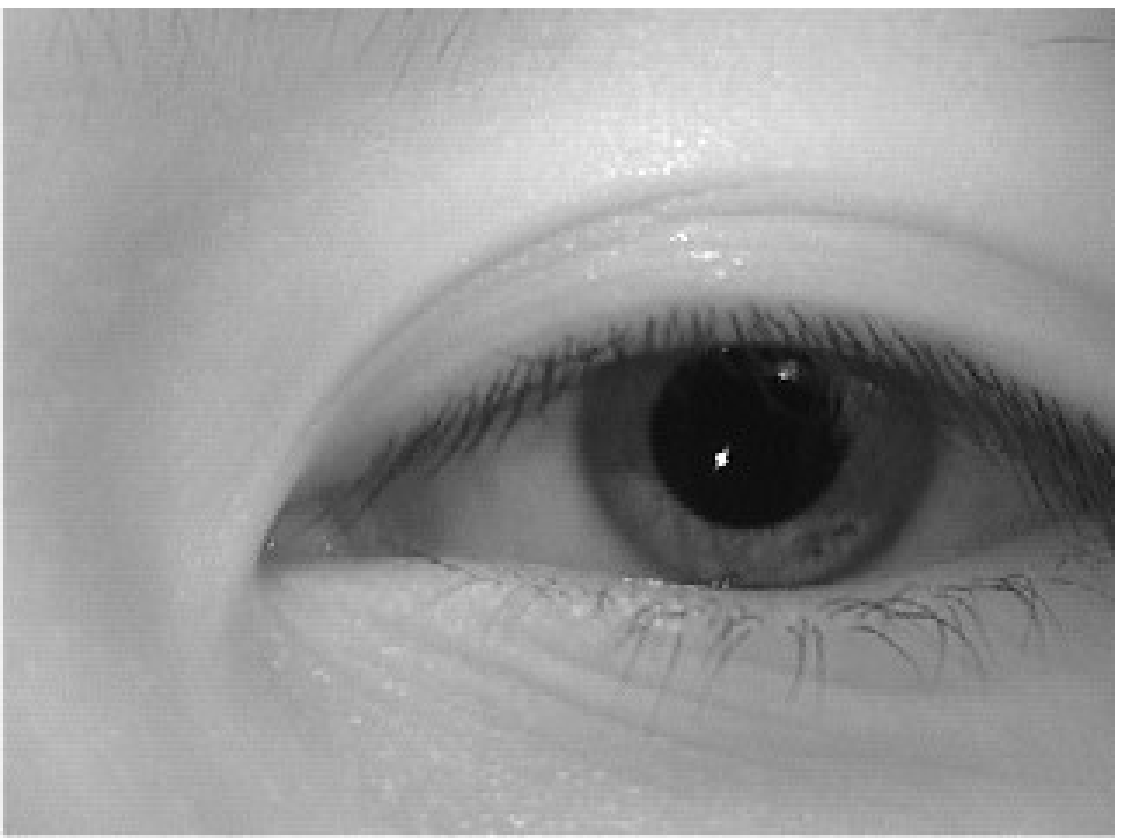}}\\
        \subfigure[] {\label{compasiron:-c}\includegraphics[scale=0.3]{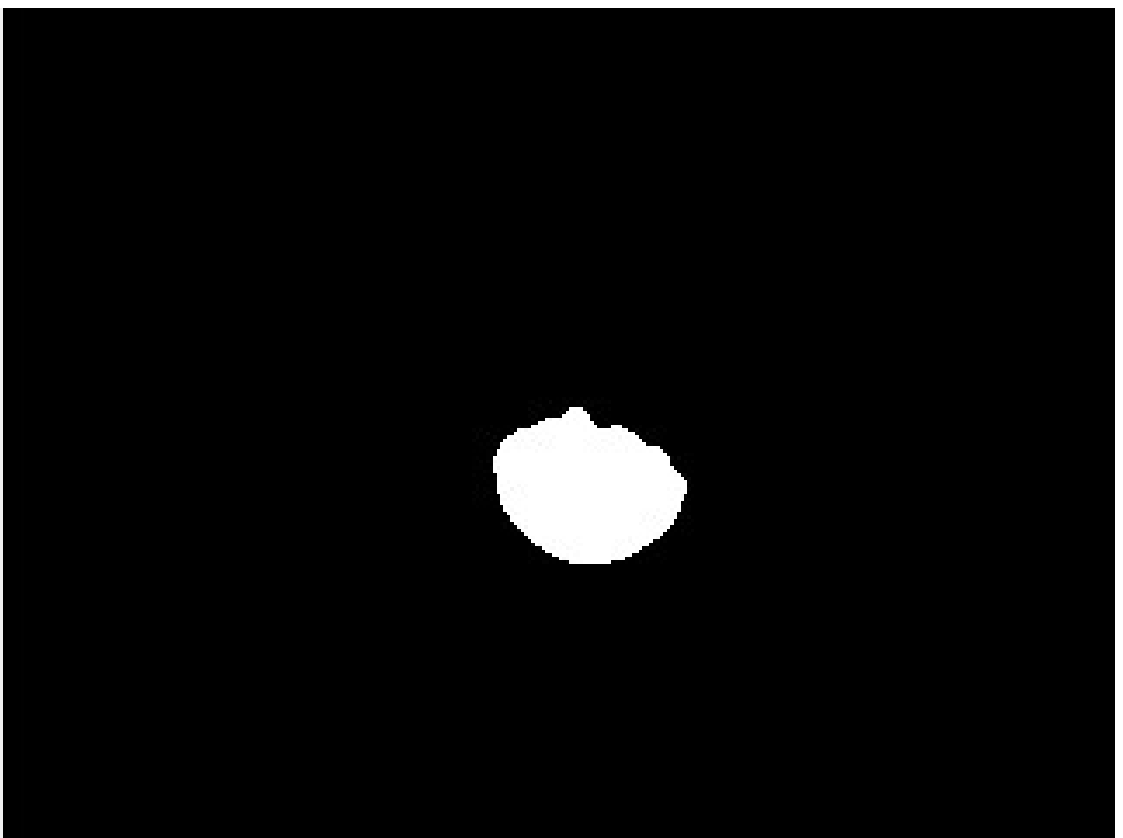}}
        \subfigure[] {\label{compasiron:-d}\includegraphics[scale=0.3]{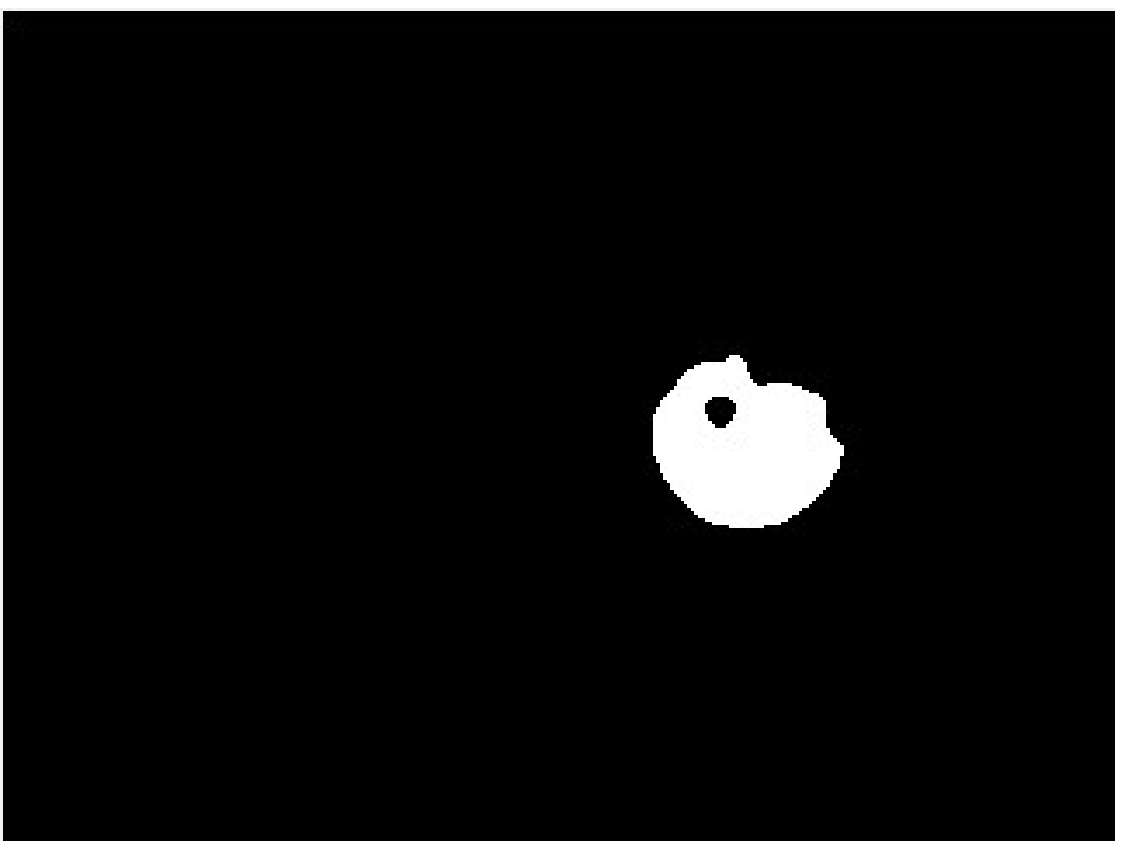}}
         \subfigure[] {\label{compasiron:-e}\includegraphics[scale=0.3]{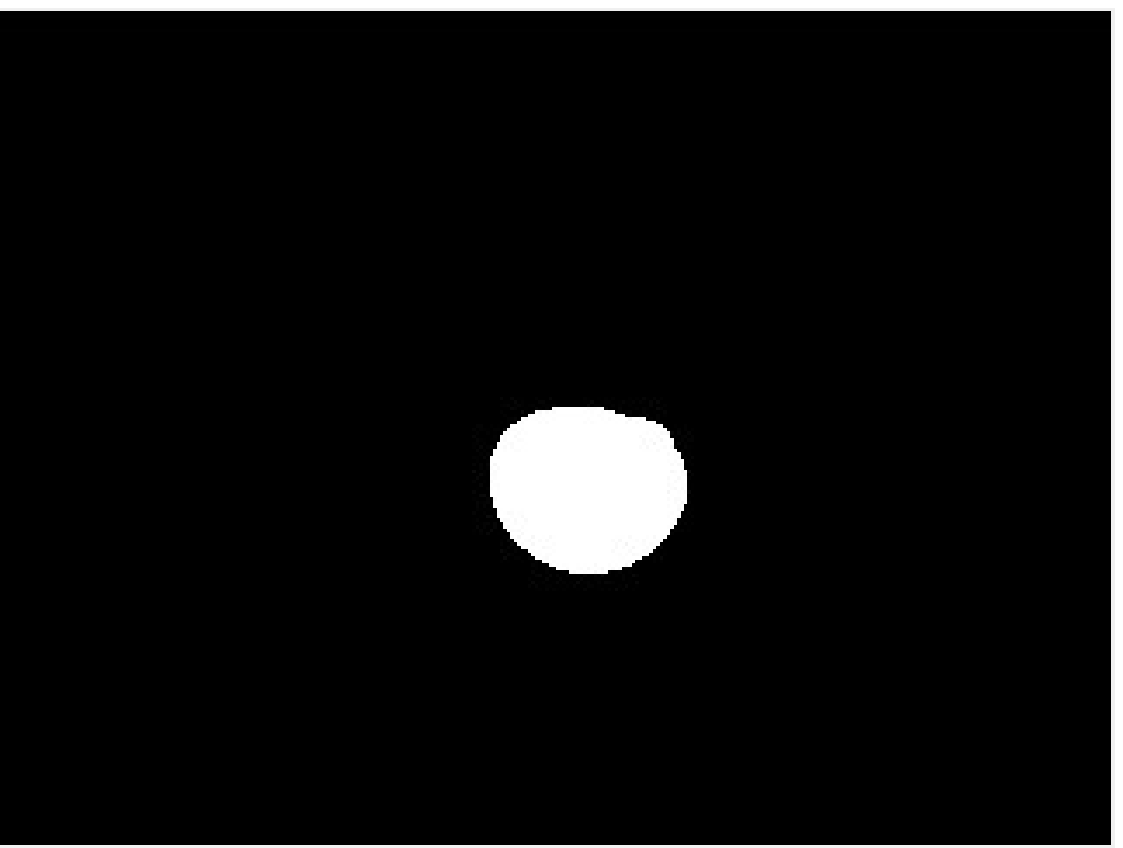}}
 		\subfigure[] {\label{compasiron:-f}\includegraphics[scale=0.3]{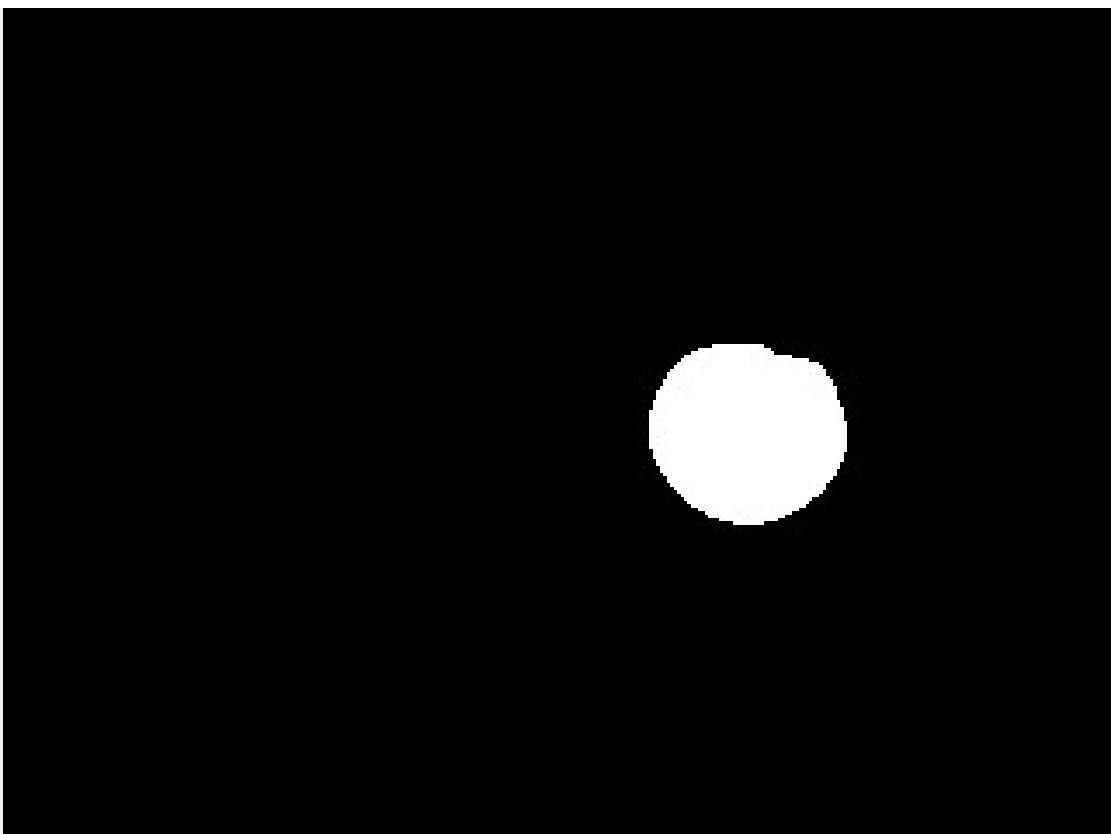}}\\
        \subfigure[] {\label{compasiron:-g}\includegraphics[scale=0.3]{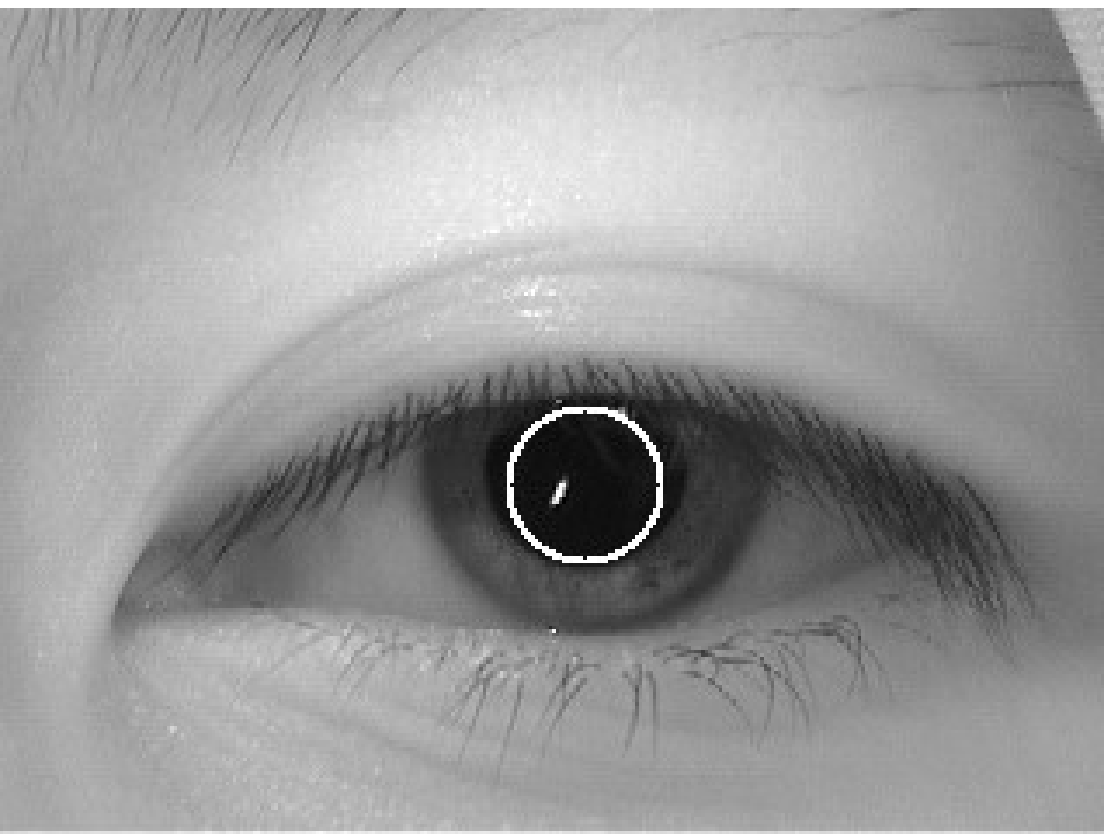}}
        \subfigure[] {\label{compasiron:-h}\includegraphics[scale=0.3]{FIGURES/p_circle1}}
        \subfigure[] {\label{compasiron:-i}\includegraphics[scale=0.3]{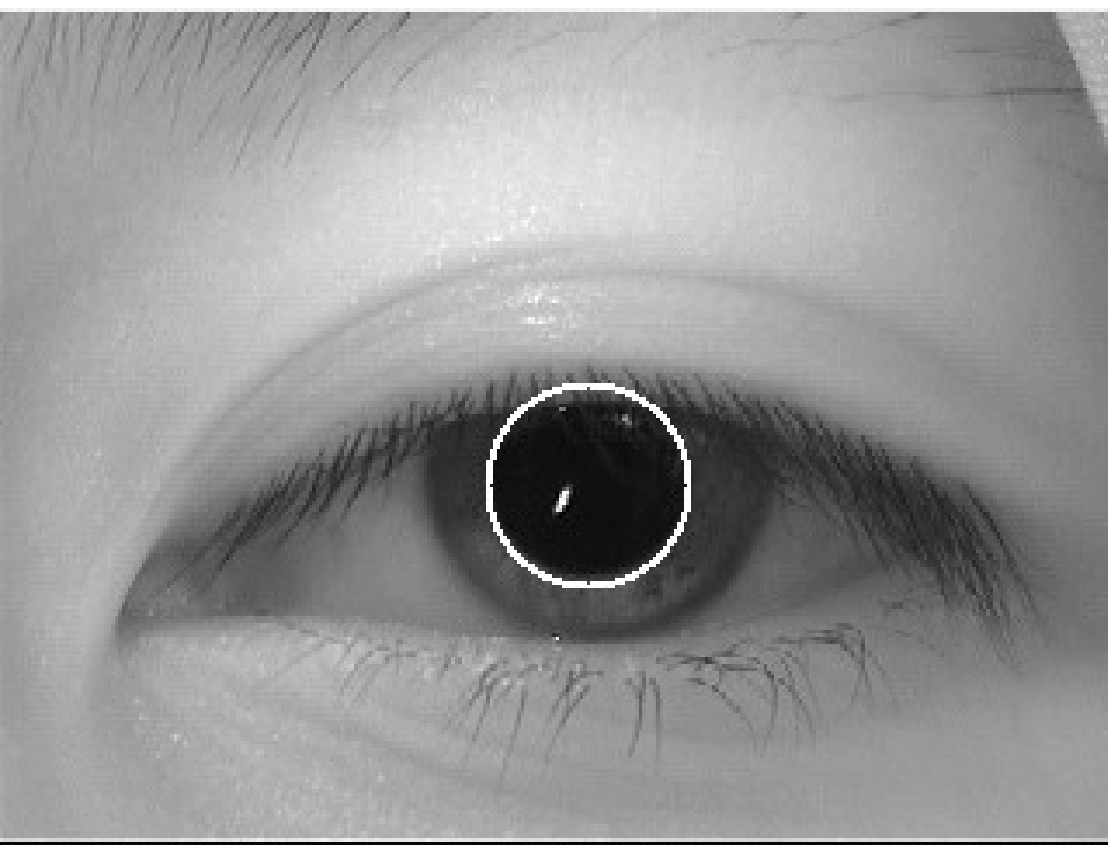}}
        \subfigure[] {\label{compasiron:-j}\includegraphics[scale=0.3]{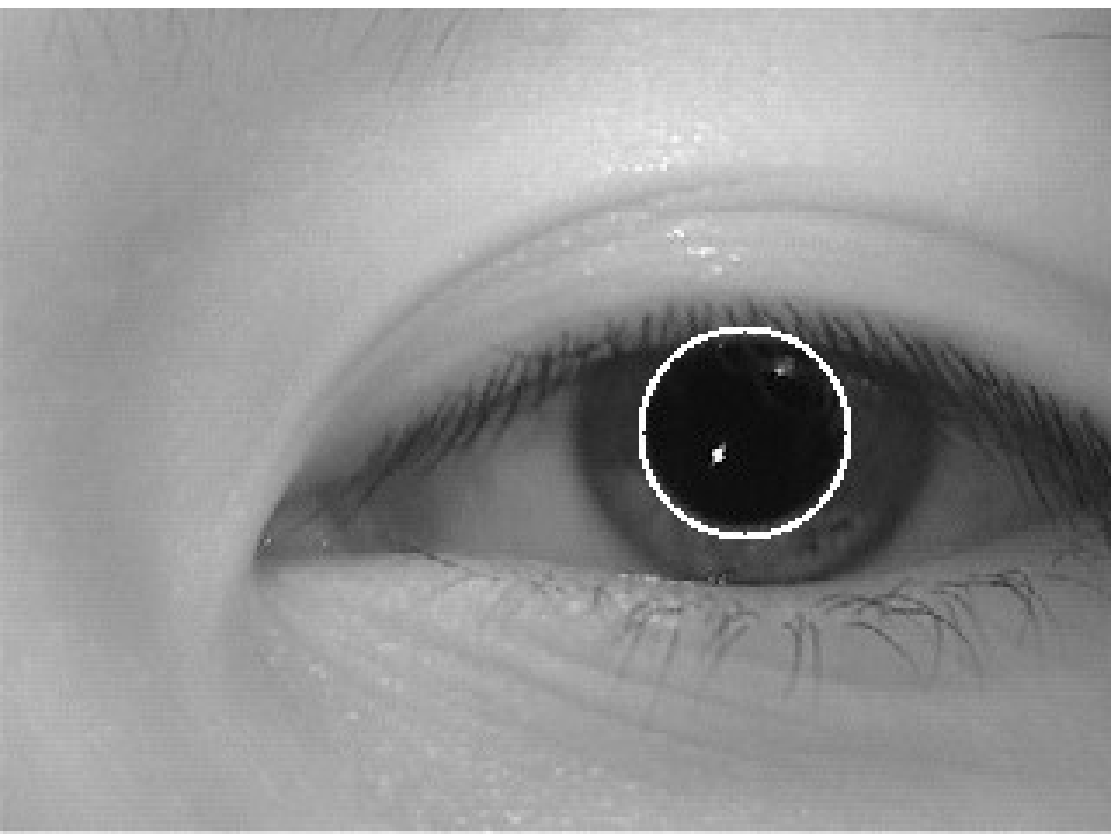}}
 	\end{center}
 	\caption{Comparison of \cite{khan2011automatic,ibrahim2012iris} with proposed method. (a) and (b) are two noisy images from the MMU v1 database. (c) and (d) show the pupil extracted by \cite{ibrahim2012iris}. (e) and (f) show the pupil extracted by the proposed method. (g) and (h) show the normalised pupil using \cite{ibrahim2012iris}. (i) and (j) show the normalised pupil by the proposed method. }
 	\label{comparison}
 \end{figure*}

\begin{figure}[htbp]
 	\begin{center}
 		\subfigure[] {\label{Med:-a}\includegraphics[scale=0.375]{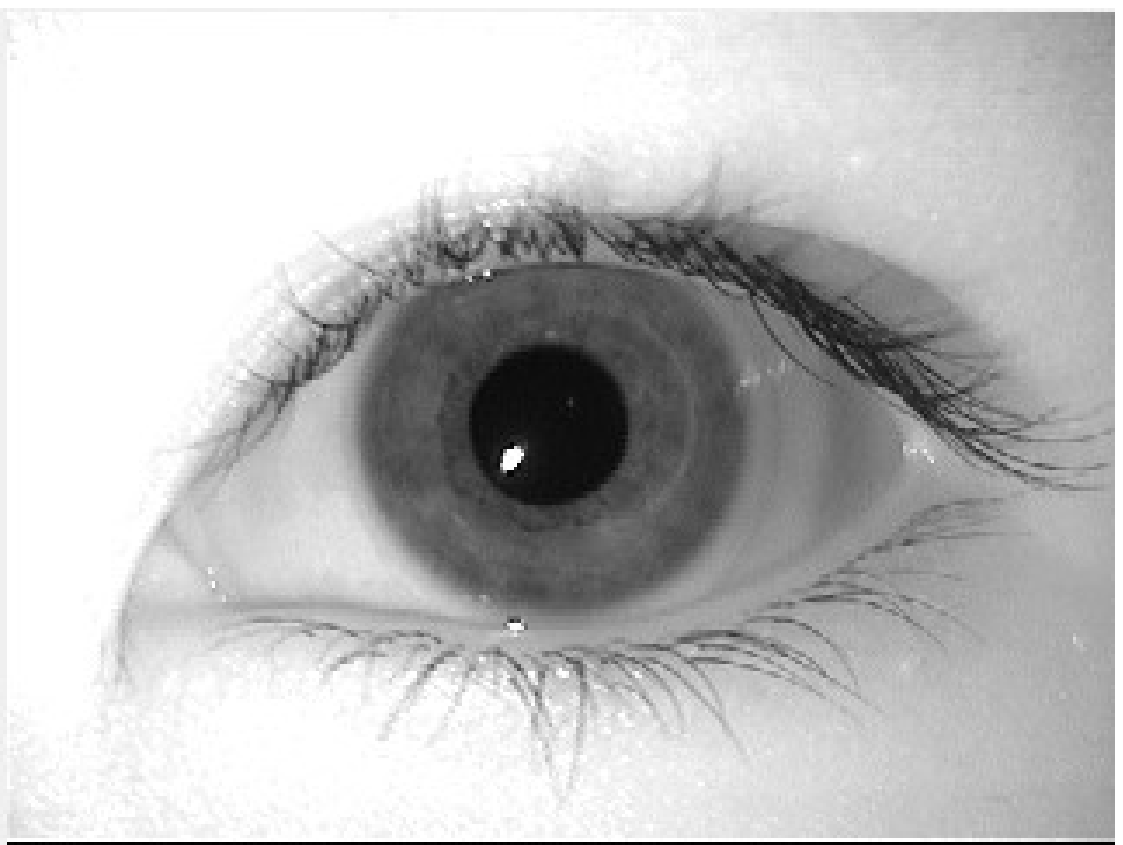}}
 		\subfigure[] {\label{Med:-b}\includegraphics[scale=0.5]{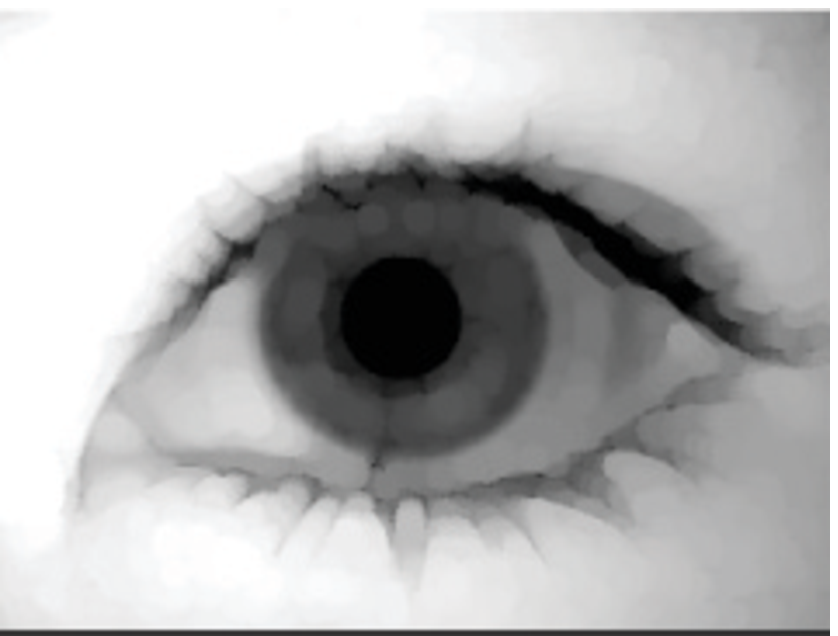}}
 	\end{center}
 	\caption{Preprocessing: (a) Sample input image.  (b) Image filtered with morphological \emph{opening} with a disc structuring element of radius 5 pixels. }
 	\label{fig:pupil}
 \end{figure}

\begin{figure*}[htbp]
 	\begin{center}
 		\subfigure[] {\label{Tri:-a}\includegraphics[scale=0.5]{FIGURES/sampleimg}}\\
 		\subfigure[] {\label{Tri:-b}\includegraphics[scale=1.1]{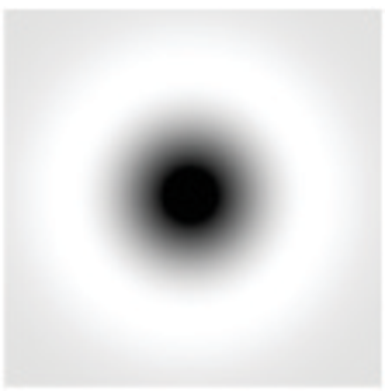}}
 		\subfigure[] {\label{Tri:-c}\includegraphics[scale=0.67]{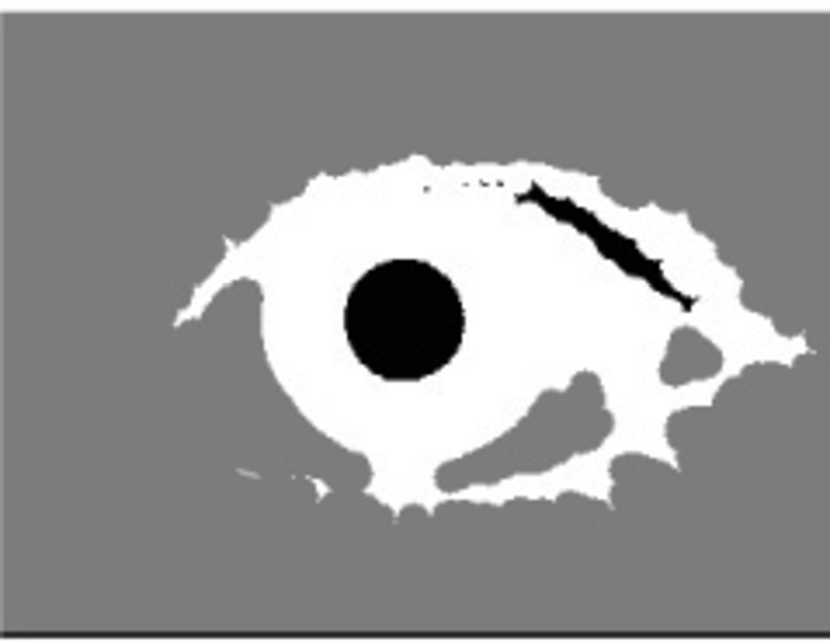}}
 	\end{center}
 	\caption{Median Filtering: (a) Sample input image.  (b) 2D representation of a LoG filter. (c) image (a) converted to a tri-level image }
 	\label{Tri_level_image}
 \end{figure*}

\begin{figure*}[htbp]
 	\begin{center}
 		\subfigure[] {\label{LoG:-a}\includegraphics[scale=0.5]{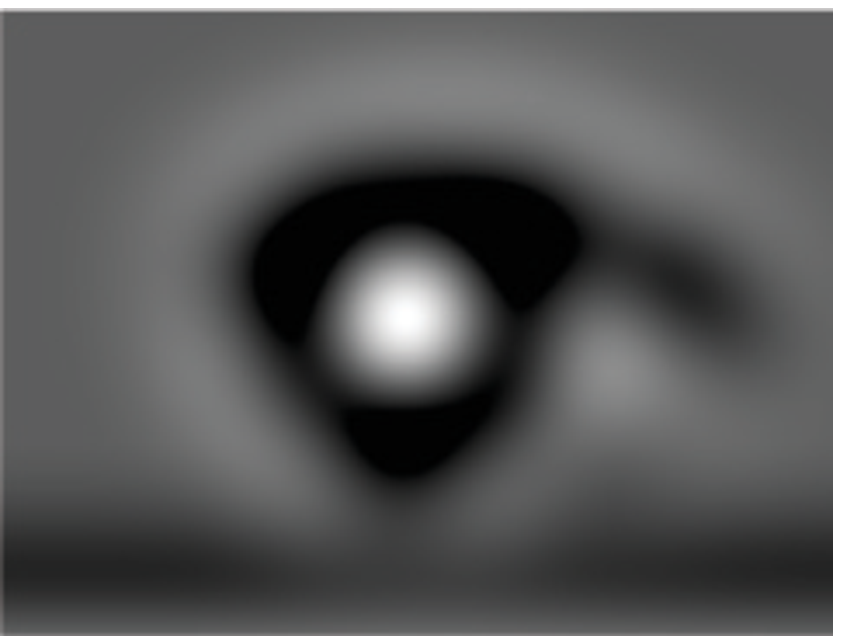}}
        \subfigure[] {\label{LoG:-c}\includegraphics[scale=0.5]{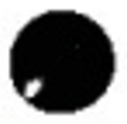}}
        \subfigure[] {\label{LoG:-d}\includegraphics[scale=0.5]{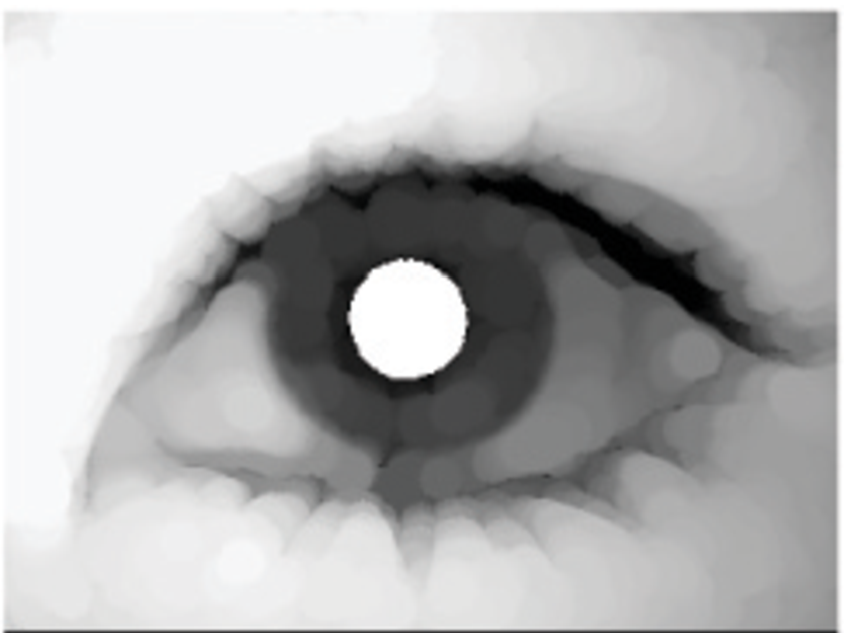}}
 	\end{center}
 	\caption{LoG filtering: (a) LoG filtered image. (b) seed image created from threshold image. (c) Segmented pupil by proposed method }
 	\label{LoG_img}
 \end{figure*}

There are many methods reported in the literature for detecting circular objects in an image. Intensity-based thresholding can be treated as blob detection, in which the pupil is treated as a circular black region on a bright background. Such techniques assume that the gray-level intensity of the pupil in an eye image is less than for any other region (e.g., iris, sclera, and skin parts). Some researchers use this assumption to localise coarse pixels in the pupil region using a histogram or threshold and then use gradient-based techniques to segment the boundaries of the iris \cite{khan2011automatic, ibrahim2012iris, Basit2007}. These techniques may not work for an eye image containing other low-intensity regions. When the pupil is shaded by eyelashes, thresholding fails to locate the true centre and radius of the pupil region. For example,  Fig. \ref{compasiron:-e} and \ref{compasiron:-f} clearly show that \cite{ibrahim2012iris} fails to get a proper pupil region because of noise. To overcome this, \cite{Jan2014} proposed some solutions that make the implementation computationally expensive. Another disadvantage of \cite{Jan2014} is the use of iterative processes that limit its real-time implementation.\\
To provide a robust, computationally light and non-iterative solution for finding the disc, we adopted a scheme that combines both a shape-based detector and an intensity-based threshold. We treated the pupil as a blob and tune a Laplacian of Gaussian (LoG) \cite{Lindeberg1998} operator to detect the edges or blobs at a particular scale. It is based on filtering an image with a Gaussian of particular standard deviation $\sigma$, also known as the scale factor. The 2-D LoG function with Gaussian standard deviation $\sigma$ has the form:
 \begin{equation}\label{PSEq1}
h\left( {x,y;\sigma } \right) =  - \frac{1}{{\pi {\sigma ^4}}}\left[ {1 - \frac{{{x^2} + {y^2}}}{{2{\sigma ^2}}}} \right]{e^{ - \frac{{\left( {{x^2} + {y^2}} \right)}}{{2{\sigma ^2}}}}}
 \end{equation}
 The scale normalised version of the LoG filtered defined in Eq. \ref{PSEq1} is modified as \cite{Lindeberg1998}:
 \begin{equation}\label{NSEq}
\begin{array}{l}
{h_{SN}}\left( {x,y;\sigma } \right) = {\sigma ^2} \times h\left( {x,y;\sigma } \right)\\
\,\,\,\,\,\,\,\,\,\,\,\,\,\,\,\,\,\,\,\,\,\,\,\,\,\, =  - \frac{1}{{\pi {\sigma ^2}}}\left[ {1 - \frac{{{x^2} + {y^2}}}{{2{\sigma ^2}}}} \right]{e^{ - \frac{{\left( {{x^2} + {y^2}} \right)}}{{2{\sigma ^2}}}}}
\end{array}
 \end{equation}
The selection of $\sigma$ depends on the blob size. A Gaussian has several advantages that facilitate blob detection. First, the Gaussian is separable; that makes it computationally efficient. Second, the Gaussian is smooth and localised in both spatial and frequency domains. This smoothing provides a good compromise in terms of suppressing false edges. The LoG acts as a bandpass filter because of its differential and smoothing behaviour. As the Laplacian is a linear operator, Gaussian filtering followed by differential is the same as filtering with a Laplacian of Gaussian.\\
 \indent To provide a robust and computationally light solution for finding the disc, we adopt the following strategy. An iris image $I\left( {x,y} \right)$ from the MMU-V1 database is taken as sample, as shown in Fig. \ref{Med:-a}. The image contains a black disc in the centre of the eye with small clumps of undesirable foreground pixels, e.g. salt noise. Though a median-based operator can be used to tackle such high-frequency noise, it can be computationally expensive. Therefore, it can be replaced with a morphological opening that does a similar job with fewer resources. The smoothed image is shown in Fig. \ref{Med:-b}.\\
The next processing step is to apply the LoG filter. The LoG filter is an anisotropic filter that has been used effectively in the past to detect blobs \cite{Lindberg_90}. For iris segmentation, the pupil circular region can be taken as a black blob on a white background. To facilitate the LoG application, the iris image is first converted to a tri-level image. In the tri-level image, the pupil is represented by black intensity, the outer circular region around the pupil with white intensity, and the rest of the image as grey intensity. This particular swapping of the white and grey level regions facilitates the application of the LoG filter, by giving a larger contrast between the inner and outer circles as shown in Fig. \ref{Tri:-b}.\\
  \indent Conversion to a tri-level image requires two thresholds: T1 and T2. The first threshold T1 is chosen as a level below which we have a large confidence of picking the pupil. The second threshold T2 is an intensity level beyond which we have a high confidence to get the rest of the image. The result is therefore given by
  \begin{equation}\label{LoGEQ}
 Tri_{img} = \left\{ \begin{array}{l}
0\,\,\,\,\,\,\,\,\,\,\,\,if\,\,\,{I_{smooth}} < T1\\
1\,\,\,\,\,\,\,\,\,\,\,\,if\,\,\,T1 < {I_{smooth}} < T2\\
0.5\,\,\,\,\,\,\,\,if\,\,\,\,\,{I_{smooth}} > T2
\end{array} \right.
  \end{equation}
 The problem of choosing the thresholds has been facilitated by preprocessing the image with morphological operator opening. This greatly reduces the stress of finding accurate thresholds $T_1$ and $T_2$. For the sake of experiments, we choose T1=0.2 and T2=0.5 for a scaled IRIS image in the [0-1] range. The tri-level intensity converted image is displayed in Fig. \ref{Tri:-c}. The figure shows the effectiveness of tri-level scaling. \\
A LoG filter given in Eq. \ref{PSEq1} is applied on ${Tri_{img}}$ as
 \begin{equation}\label{EQLoG}
 {I_{LoG}} = {Tri_{img}}*h\left( {x,y,\sigma } \right),
 \end{equation}
  where $\sigma=R$ is the average pupil radius of a particular database. By applying this LoG with such a coarse scale, the output images ${I_{LoG}}$ possesses strong contours due to the heavy smoothing, as shown in Fig. \ref{LoG:-a}. The LoG filtering provides the maximum response in the pupil region. Now a mask is created that corresponds to the pixel of ${I_{LoG}}$  with maximum response as
  \begin{equation}\label{PSEq3}
{Ig_{mask}} = \left\{ \begin{array}{l}
1,\,\,\,\,\,\,if\,\,\,\,{I_{LoG}} > {\lambda _a}\\
0,\,\,\,\,\,\,otherwise
\end{array} \right..
  \end{equation}
where ${\lambda _a} $ is set to some value above the mid-grey level, such as 0.6 for $I_{LoG}$ scaled in [0-1] range. Fig. \ref{LoG:-b} shows the mask image that is used for generating the first seed image. This mask image ${Ig_{mask}}$ is then multiplied pixel-wise by the ${I_{smooth}}$ image to get the first seed image as
\begin{equation}\label{PSEq4}
  {Ig_{seed}} = {I_{smooth}}.*{Ig_{mask}}
\end{equation}

\begin{figure*}[htbp]
 	\begin{center}
 		\subfigure[] {\label{lognoisy:-a}\includegraphics[scale=0.5]{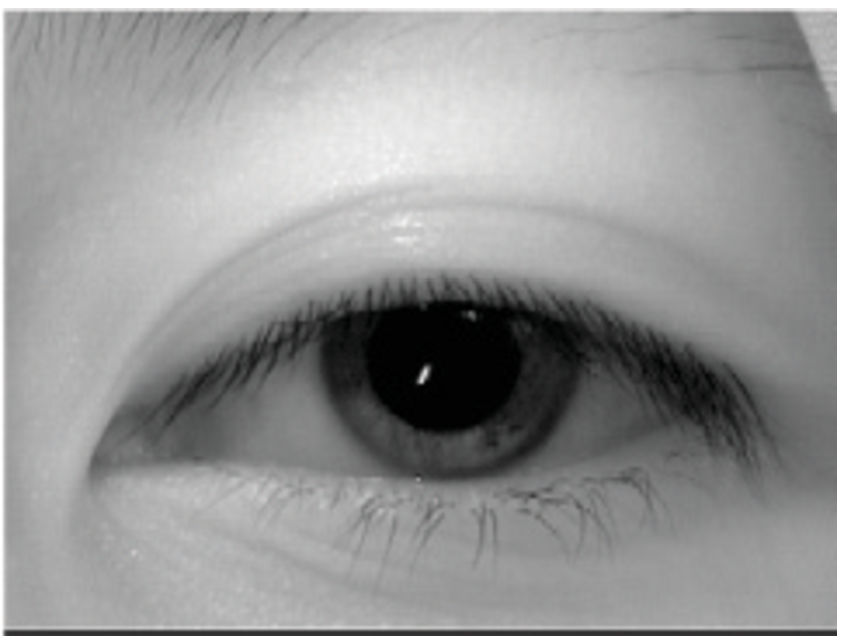}}
 		\subfigure[] {\label{lognoisy:-b}\includegraphics[scale=0.5]{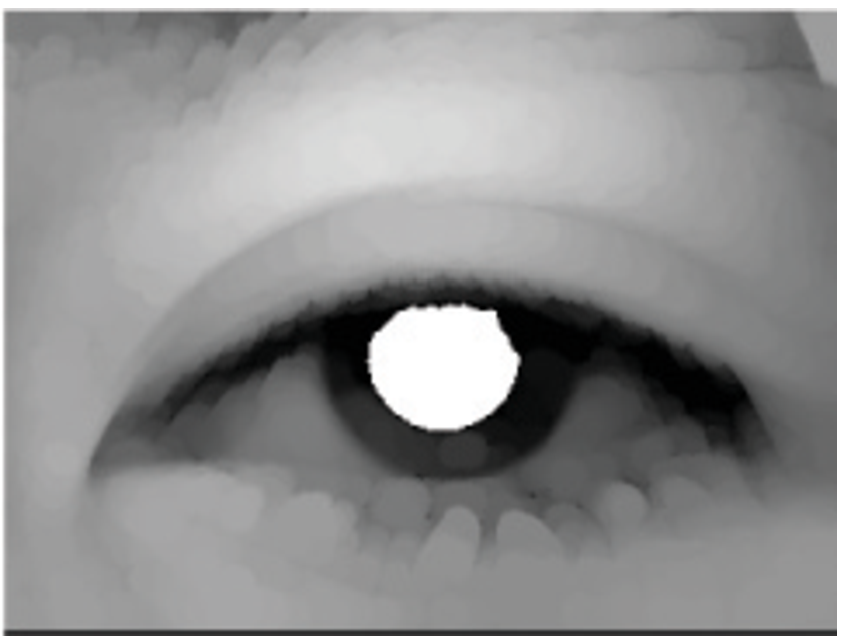}}
  		\subfigure[] {\label{lognoisy:-c}\includegraphics[scale=0.5]{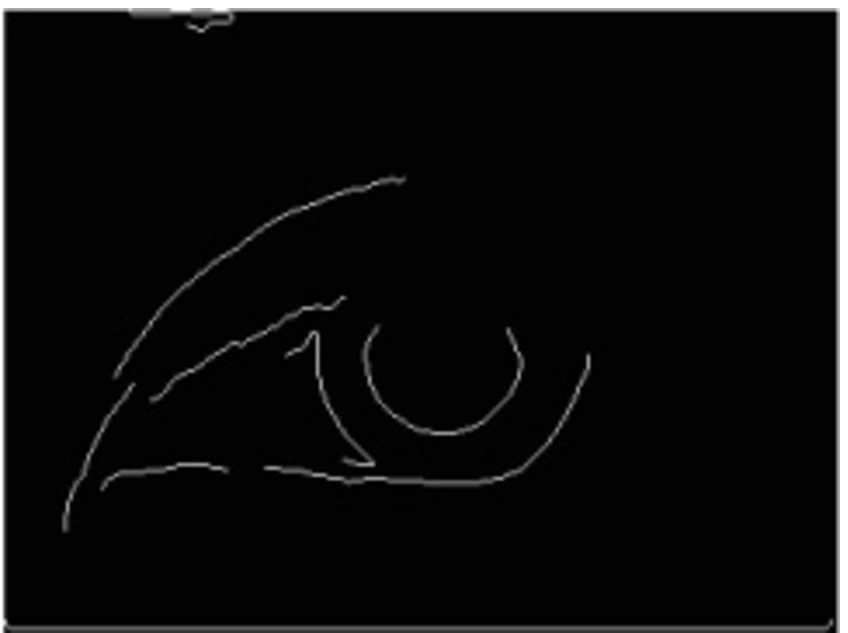}}
 	\end{center}
 	\caption{LoG filtering on noisy image: (a) Sample noisy image of MMU v1 database. (b) Detected pupil by proposed method. (c) Zero-crossing of LoG filtered image. }
 	\label{LoG_on_Noisy}
 \end{figure*}

\begin{figure*}[htbp]
 	\begin{center}
 		\subfigure[] {\label{zerocrossing:-a}\includegraphics[scale=0.5]{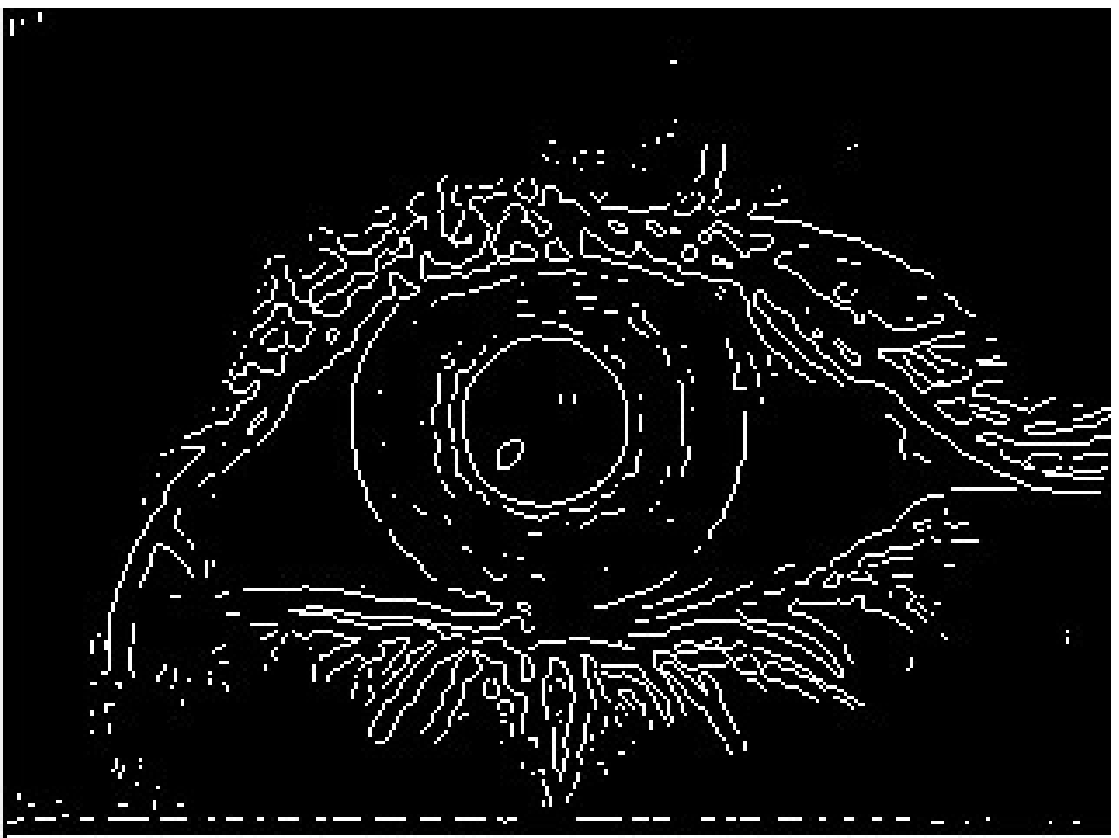}}
 		\subfigure[] {\label{zerocrossing:-b}\includegraphics[scale=0.5]{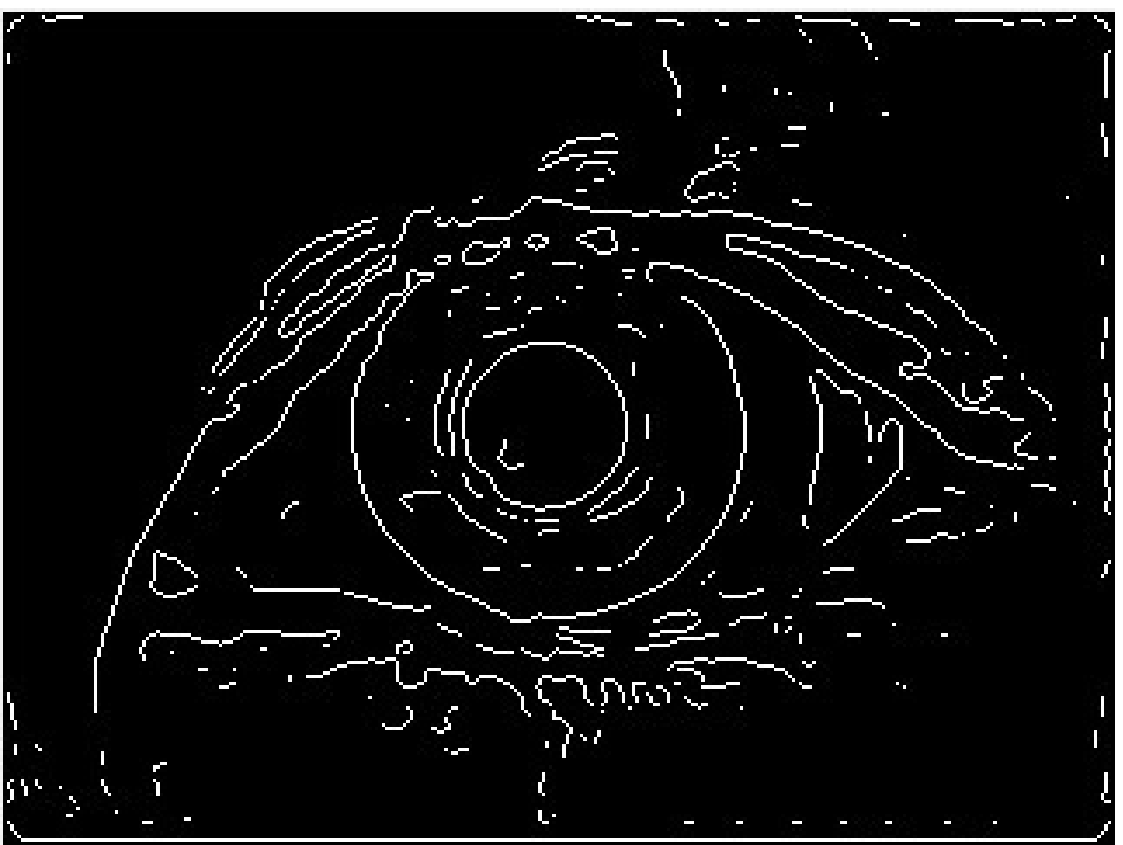}}
        \subfigure[] {\label{zerocrossing:-c}\includegraphics[scale=0.5]{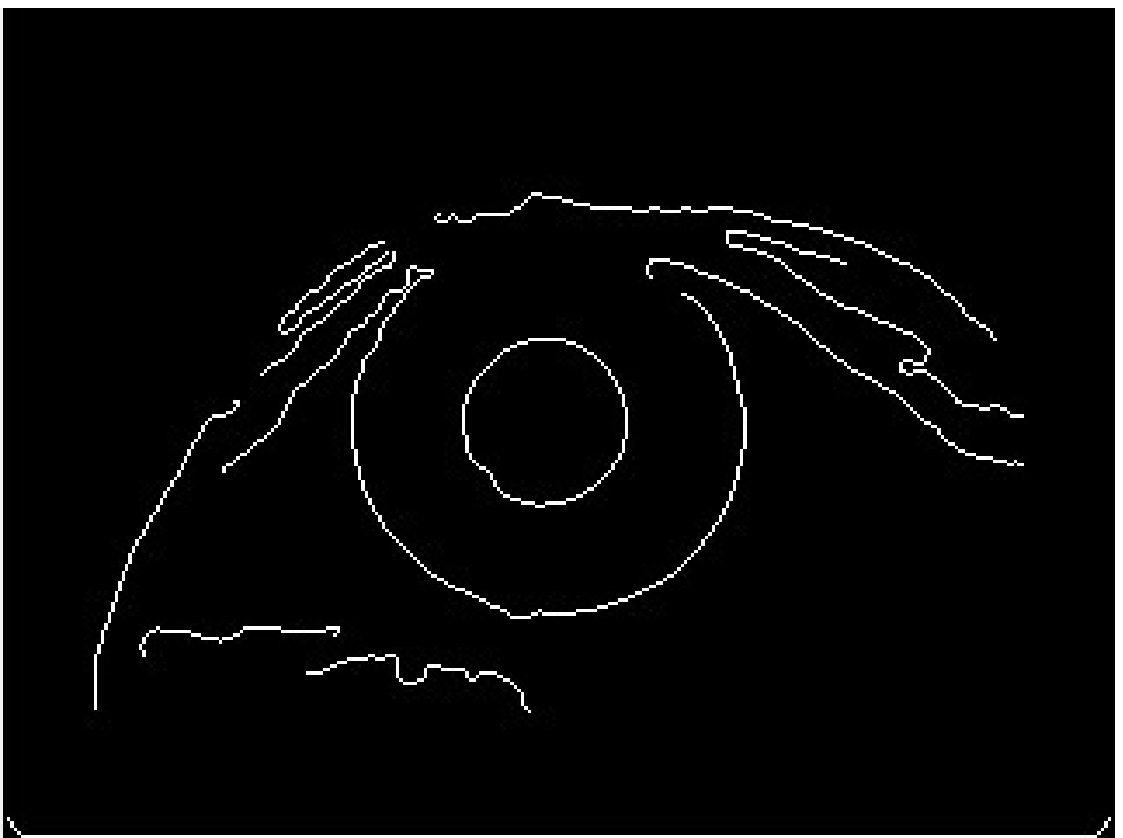}}
 	\end{center}
 	\caption{LoG zero crossing: (a) Zero crossing without morphological filtering. (b) Zero crossing with morphological filtering. (c) Zero crossing after cleaning with area property. }
 	\label{zerocrossing}
 \end{figure*}

Fig. \ref{LoG:-c} shows that this seed image contains the pupil. The next task is to choose an appropriate seed point among this seed image using centroid of the seed image. This point is used as a seed point for the region growing method \cite{Gonzalez2006}. Region growing  is a segmentation strategy that starts with a pool of only one initial seed point and then adds more  pixels to the pool that are 8-connected neighbours with similar intensity to that of the earlier seed points. The tri-level image loses important texture information, therefore, the LoG-filtered image is used only for finding seed point where the region growing is performed on the $I_{smooth}$. The initial seed point is grown to a target pool of pixels using a similarity measure where the intensity of the seed point is compared with 8 neighbours using a 5 percent rule. Fig. \ref{LoG:-d} shows a segmented pupil using the proposed method. By using LoG filtering along with region growing, most of the problems reported in \cite{khan2011automatic, ibrahim2012iris, Basit2007} for pupil segmentation are addressed. \\
\indent From Fig. \ref{compasiron:-e} and \ref{compasiron:-f} it can be observed that the pupil region reconstructed by the proposed method is an improvement on \cite{khan2011automatic, ibrahim2012iris, Basit2007}. But in extremely noisy conditions, this still can give some error in finding the accurate centre and radius of the pupil, Fig. \ref{lognoisy:-b} provides evidence of this situation. Therefore, to further strengthen the proposed method, zero-crossings of the LoG-filtered image are also obtained. The zero-crossings give the true edges of the pupillary boundary. This can help the proposed method to find more accurately the centre and radius of the pupil. \\
\indent The behaviour of the LoG zero-crossings edge detector is largely governed by the standard deviation of the Gaussian used in the LoG filter. It is common to see several spurious edges detected away from any obvious edges. To deal with spurious edges the first order differential information of the image is required. This information will provide The gradient magnitude at the zero crossing of the LoG-filtered image.  Discarding the zero crossings with a magnitude lower than a threshold will retain only the stronger edges. To implement the zero-crossings of the LoG filter, the morphological filtered  image ${I_{smooth}}$  is convolved with a LoG filter with $\sigma=2$ , having filter size $n = \left\lfloor {(3\sigma ) \times 2 + 1} \right\rfloor $. Then the zero-crossings of this LoG filtered image are obtained with a threshold of ${\lambda _c} = 0.15$ for the MMU v1 database. Fig. \ref{zerocrossing:-b} shows the result of zero-crossings. The images of the database are preprocessed with Gaussian smoothing before calculating their first order differential strength measure. A single value of ${\lambda _c}$ is appropriate to suppress spurious edges related to insignificant zero crossing points. The value of ${\lambda _c} $ is chosen after performing several experiments on the MMU-v1 database. It is observed that if the value of ${\lambda _c} $  is increased then it gives fewer edges, and in some cases it affects the pupil and iris boundary edges. On the other hand, if a smaller value is chosen then noise in the zero-crossing image increases and makes it harder to locate the pupil. From Fig. \ref{zerocrossing:-b} it can be seen that the zero-crossings image has unwanted noise that needs to be cleaned up. Connected components of the zero-crossing image are found and those with fewer fifty pixels are removed. This removes small unwanted regions without affecting the pupil and iris boundaries, as shown in Fig. \ref{zerocrossing:-c}. The seed point calculated for the region growing is used as a reference point to extract the true circle from the zero-crossings. From this reference point, the boundary of the pupil is scanned radially in the zero crossing image. From Fig. \ref{lognoisy:-c} it can be seen that the circle is broken because of eyelashes and eyelids. The zero-crossings help to find the true broken region. By using interpolation, the remaining part of the circle can easily be predicted.


\subsection{Limbic boundary localization}
\begin{figure*}[h]
 	\begin{center}
 		\subfigure[] {\label{boundart_test:-a}\includegraphics[scale=0.4]{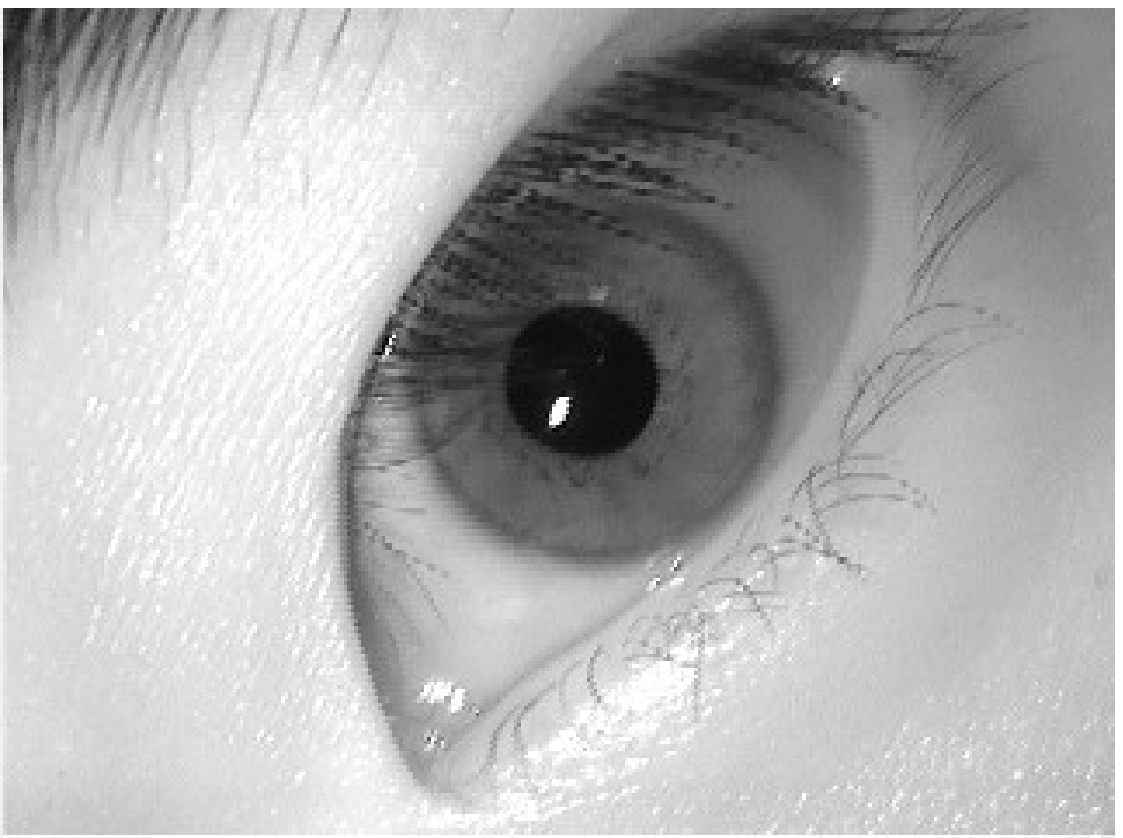}}
 		\subfigure[] {\label{boundart_test:-b}\includegraphics[scale=0.4]{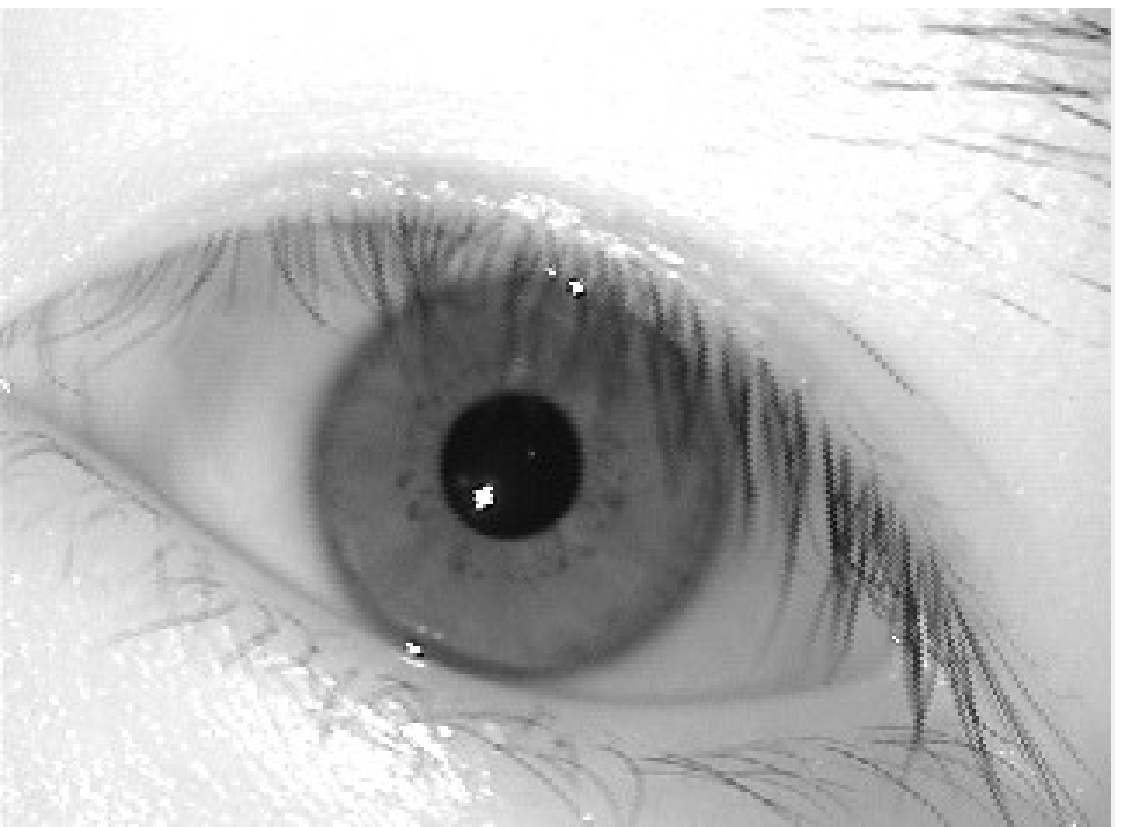}}
        \subfigure[] {\label{boundart_test:-c}\includegraphics[scale=0.4]{FIGURES/sampleimg}}\\
        \subfigure[] {\label{boundart_test:-d}\includegraphics[scale=0.4]{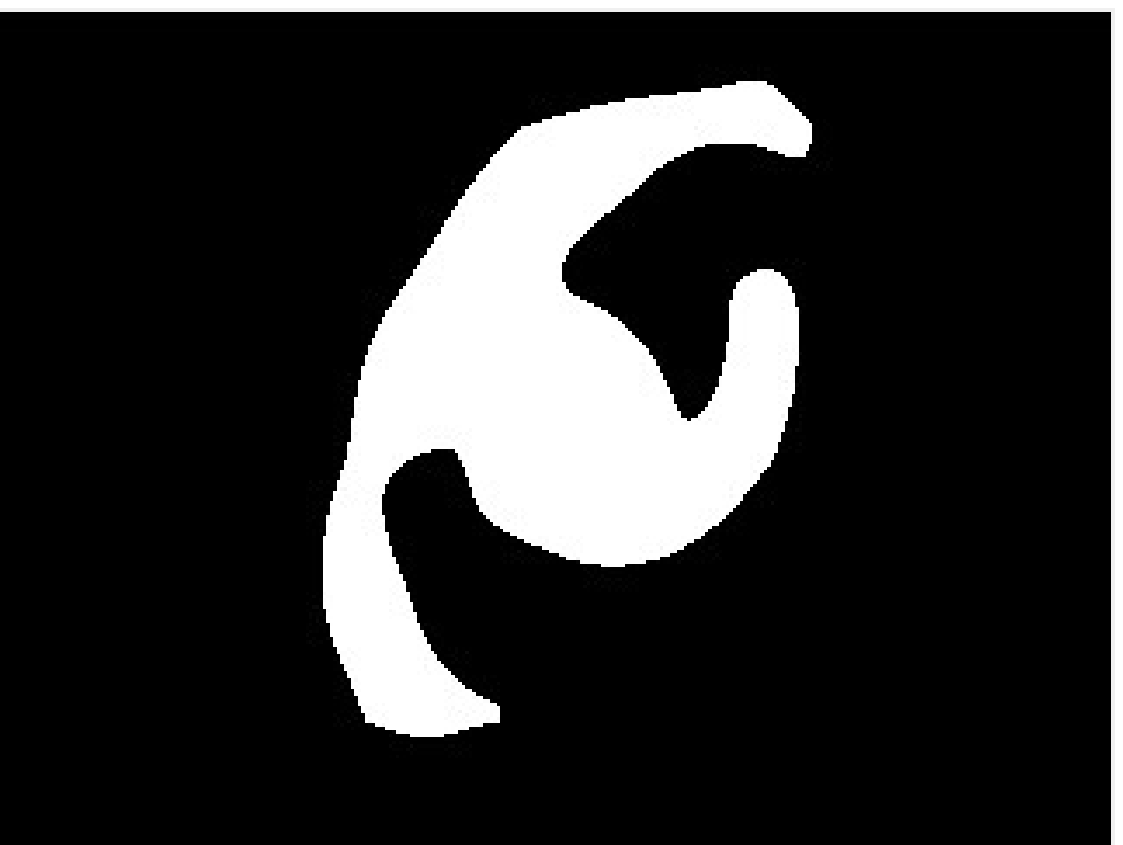}}
         \subfigure[] {\label{boundart_test:-e}\includegraphics[scale=0.4]{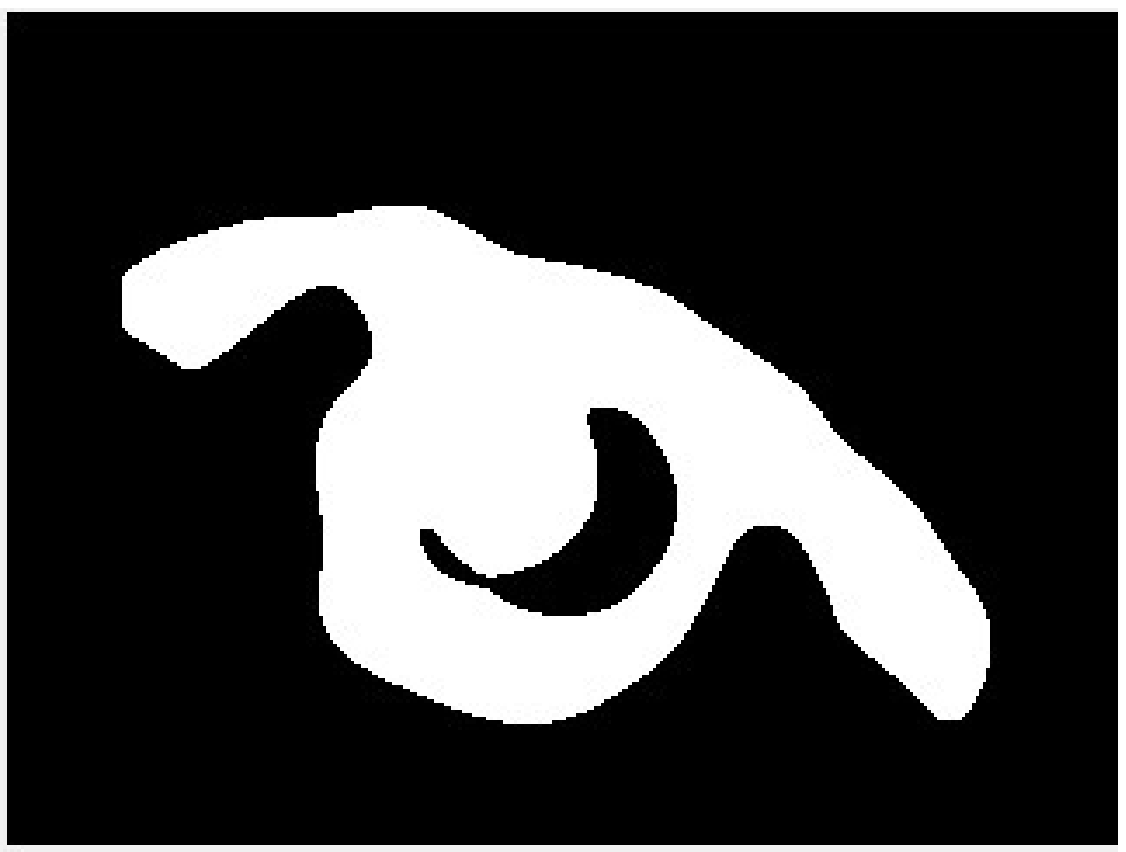}}
 		\subfigure[] {\label{boundart_test:-f}\includegraphics[scale=0.4]{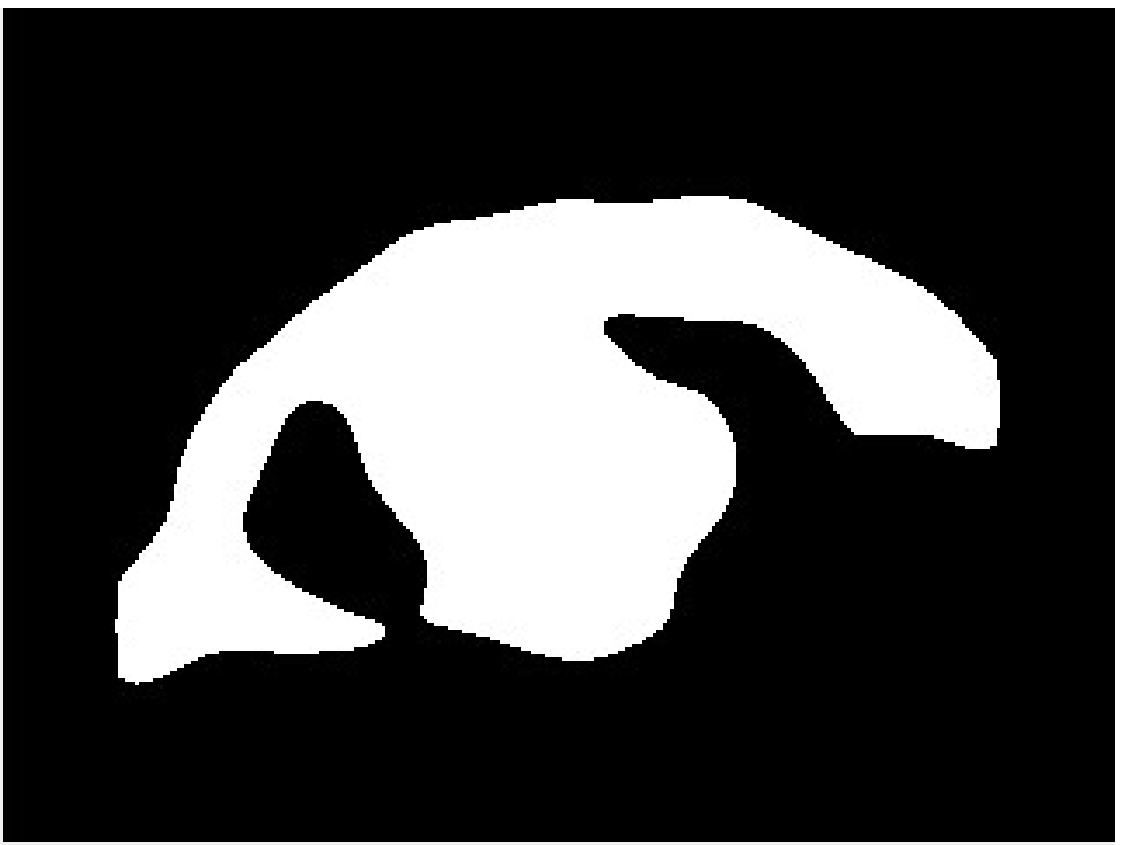}}
 	\end{center}
 	\caption{Binary Mask used to find out the true orientation of eye image. (a), (b), and (c) are three sample eye images of MMU database. (d), (e), and (f) are the binary masks of (a), (b), and (c) obtain by using LoG filtering. }
 	\label{boundart_test}
 \end{figure*}

\begin{figure*}[h]
 	\begin{center}
 		\subfigure[] {\label{OuterBoundry:-a}\includegraphics[scale=0.177]{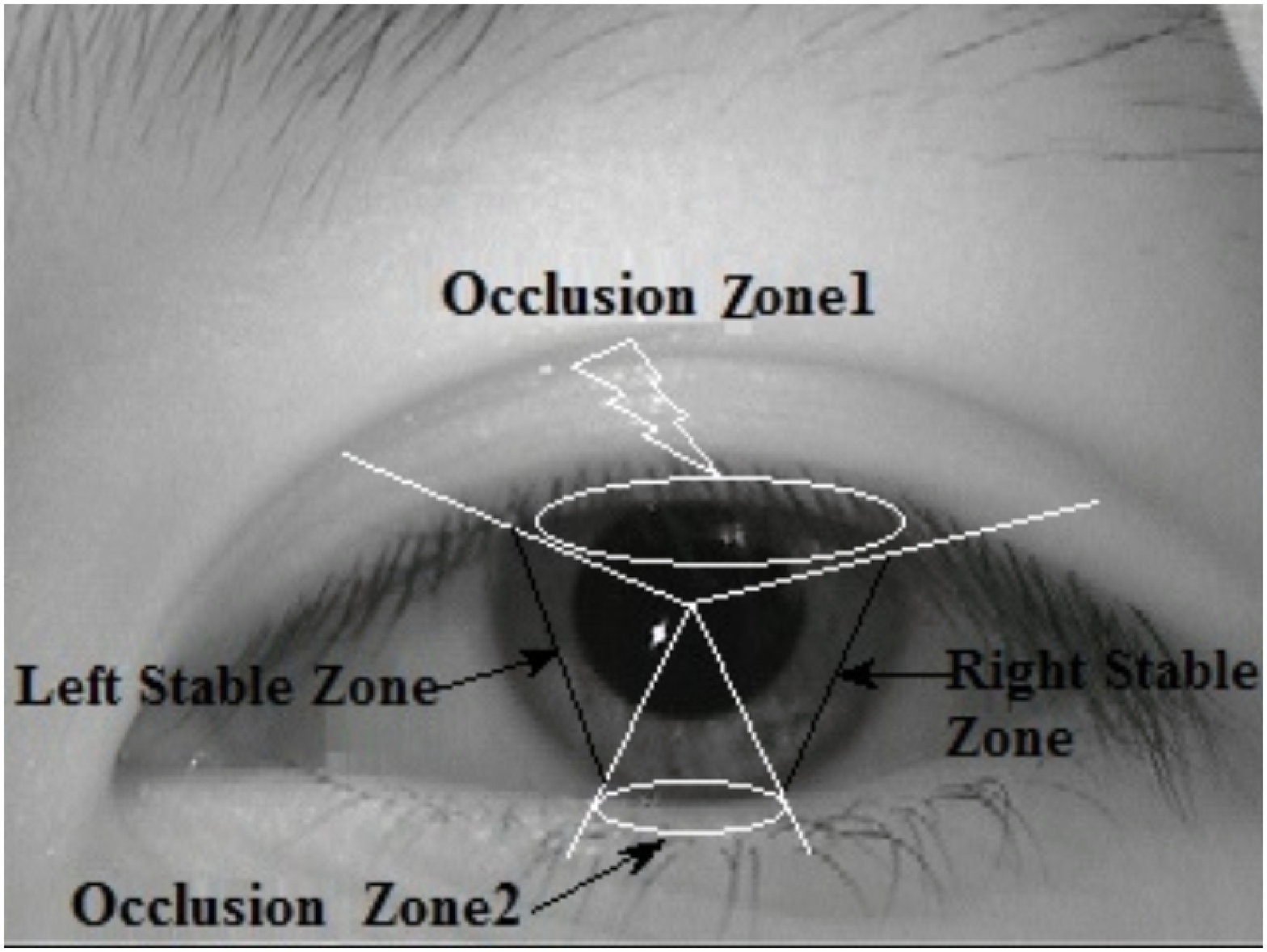}}
 		\subfigure[] {\label{OuterBoundry:-b}\includegraphics[scale=0.14]{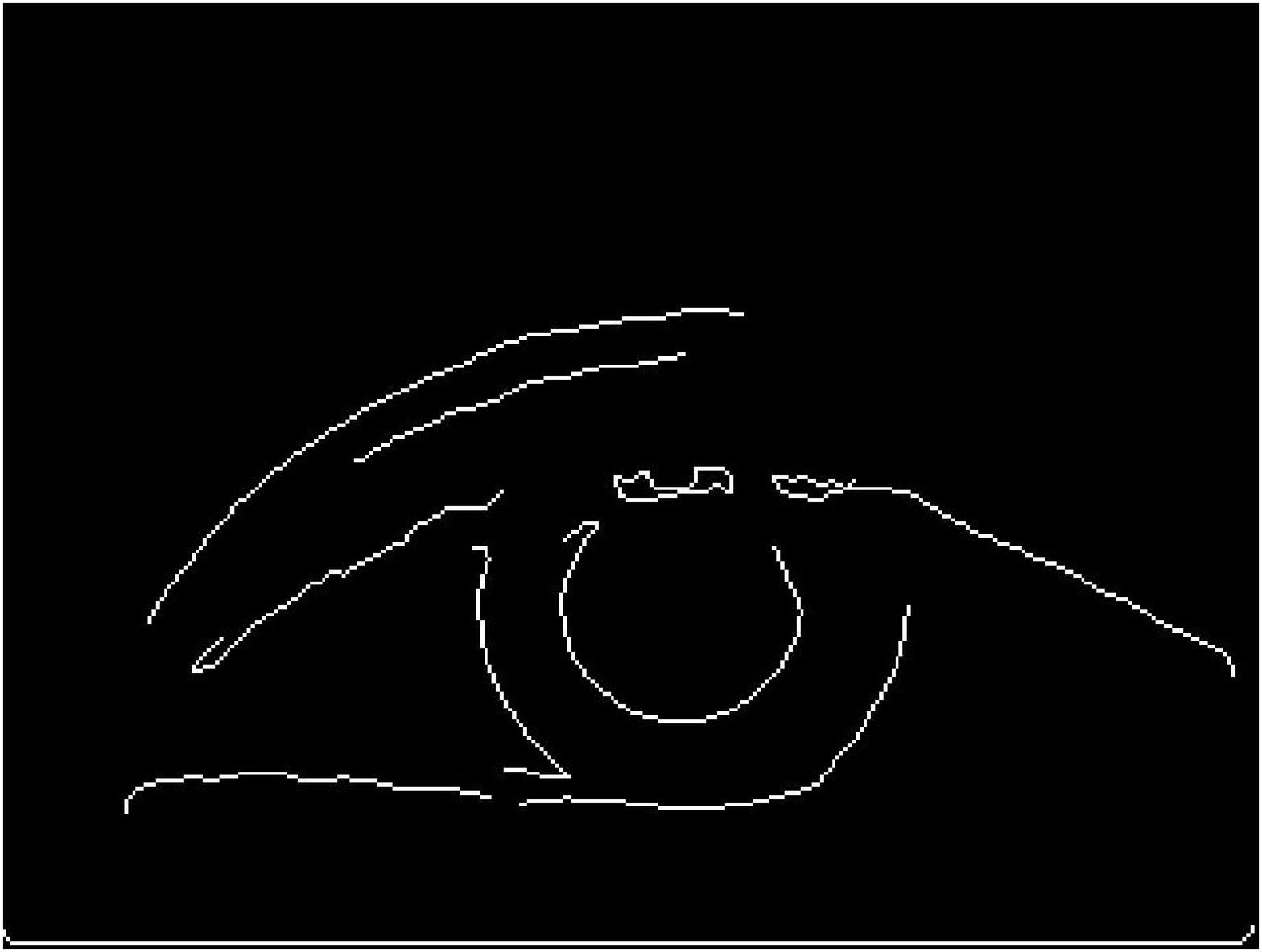}}
        \subfigure[] {\label{OuterBoundry:-c}\includegraphics[scale=0.14]{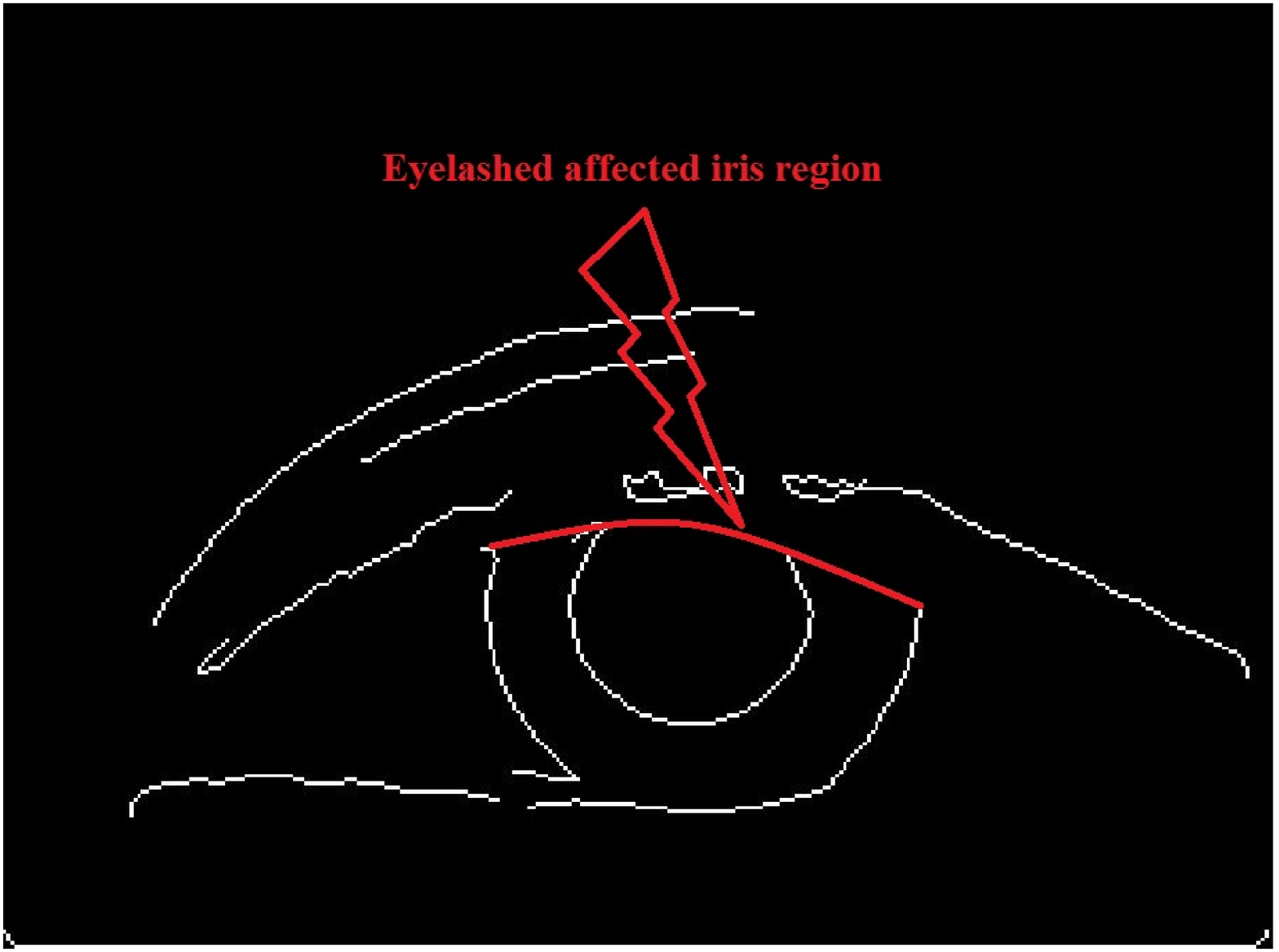}}\\
 	\end{center}
 	\caption{(a) Iris image divided into different zones, (b) Zero-crossings of LoG filtered image. (c) Detected eyelashes in iris and pupil region. }
 	\label{OuterBoundry}
 \end{figure*}

\begin{figure}
	\begin{center}
  \subfigure[] {\label{secure_region:-a}\includegraphics[scale=0.135]{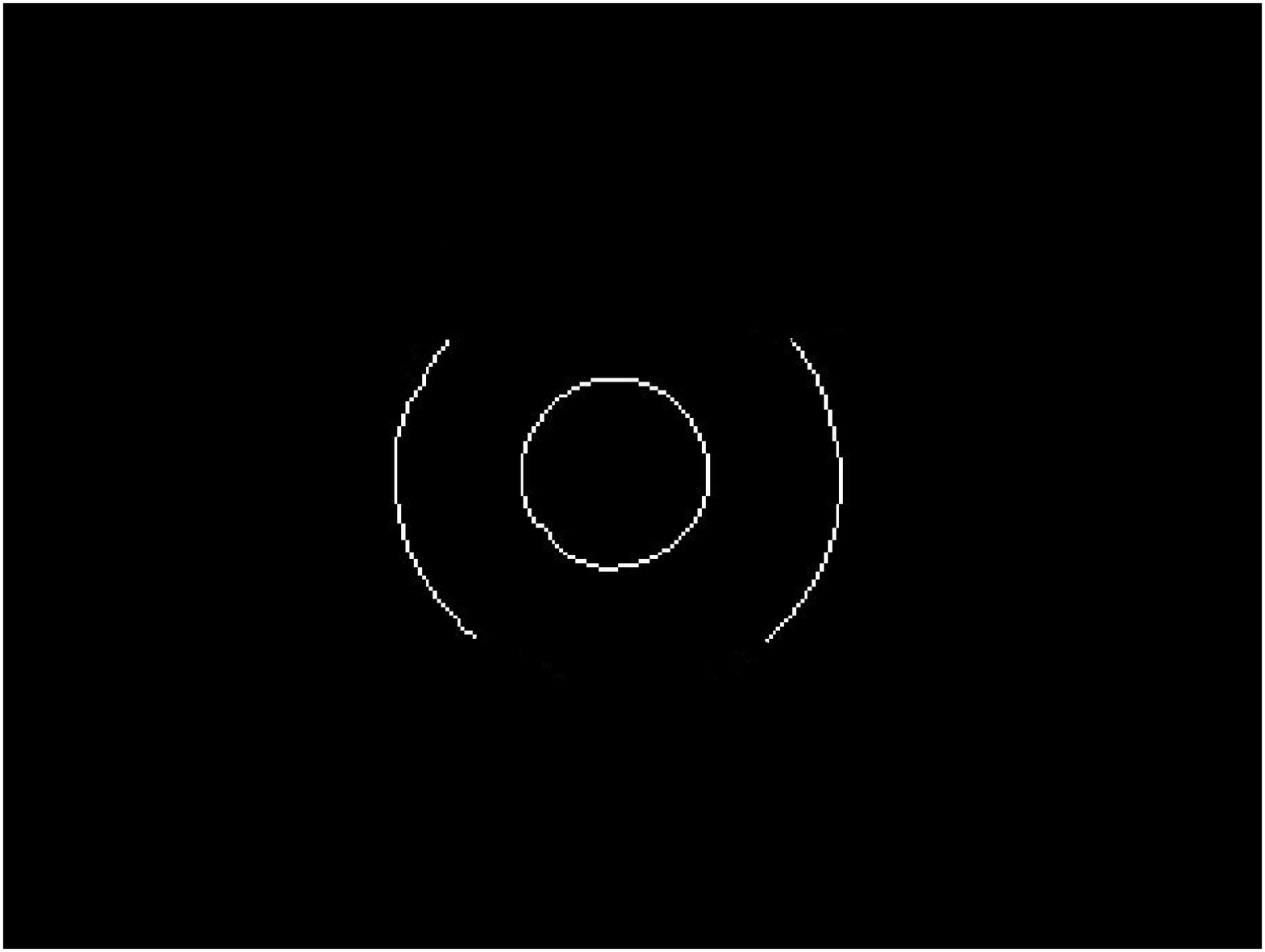}}
  \subfigure[] {\label{secure_region:-a}\includegraphics[scale=0.135]{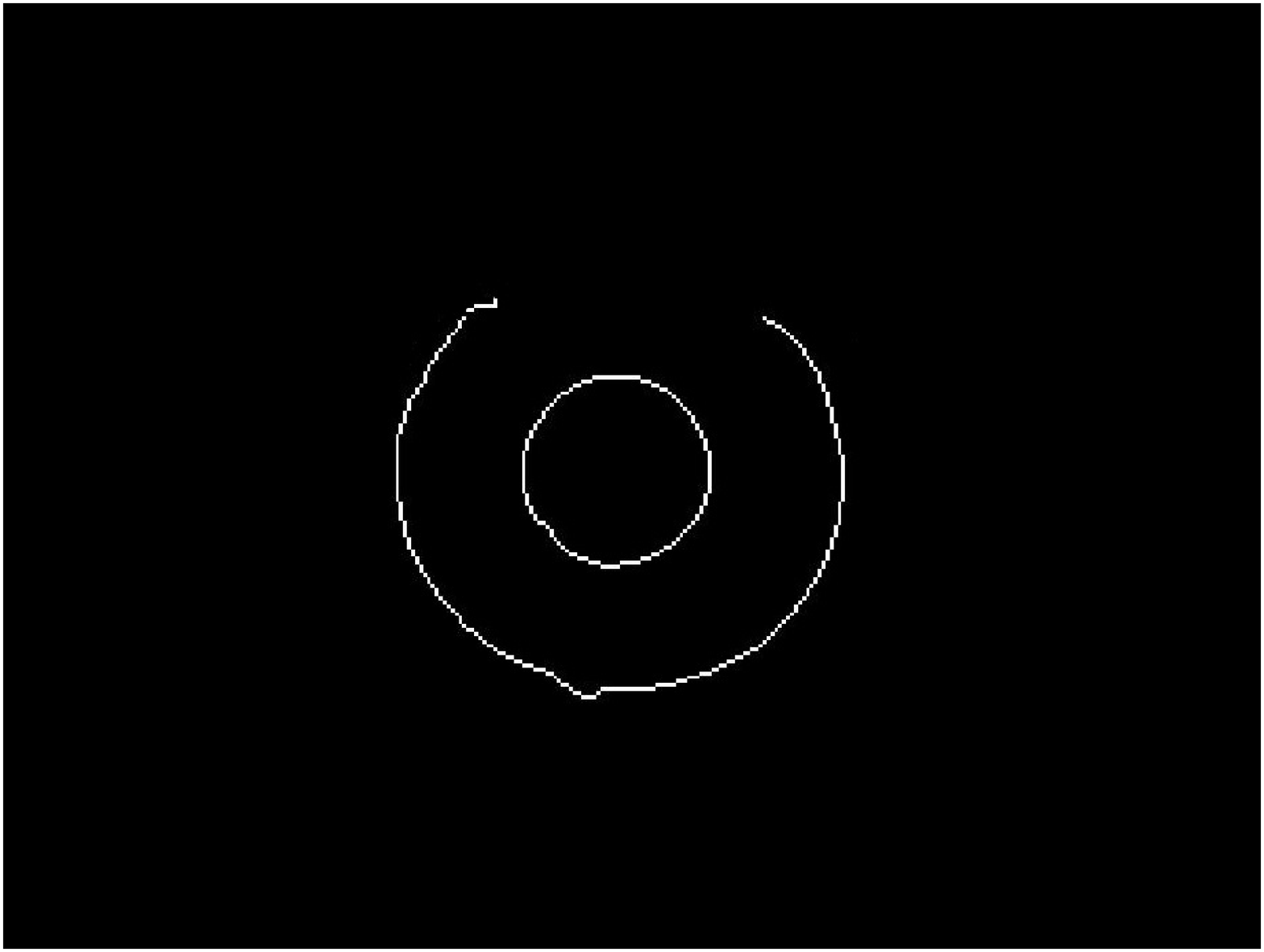}}\\
    \subfigure[] {\label{secure_region:-c}\includegraphics[scale=0.185]{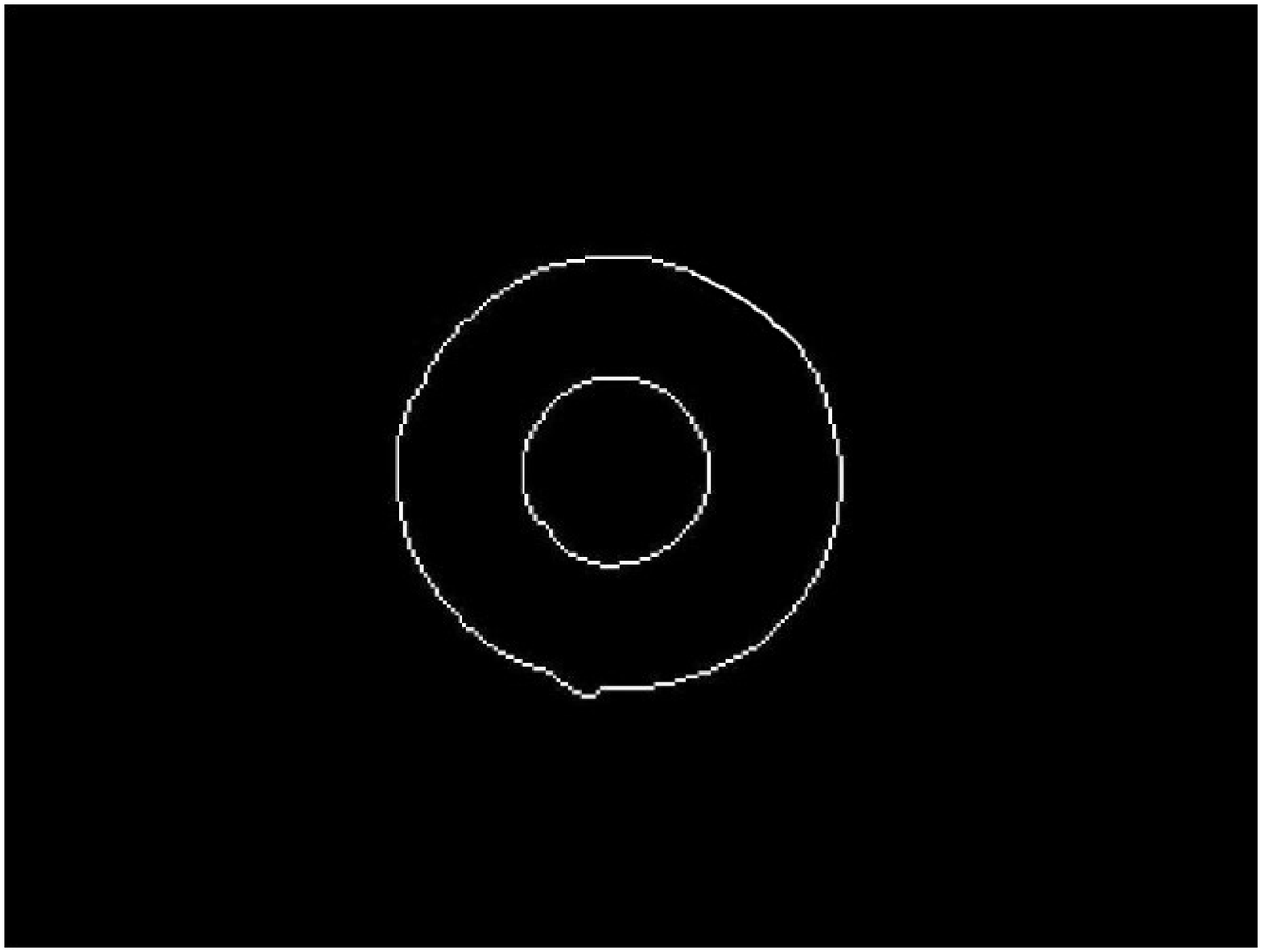}}
    \end{center}
  \caption{(a) Extracted secure region. (b) Extended search of boundaries in turbulence zones. (c) Interpolated iris.  }\label{secure_region}
\end{figure}

Limbic boundary extraction is also difficult for the following reasons: first, the eyelids and eyelashes may partially occlude the iris outer boundary. Second, the contrast between the iris and sclera regions is usually low. Lastly, as the pupil always exists within the iris region, the pupil and limbic boundaries could be assumed as two nonconcentric circles; however this assumption is not always true \cite{Bowyer2008}. Basit \cite{Basit2007} picked a horizontal line from the centre of the pupil and used a gradient to find the edges. Based on these edges, the pupil centre and the radius are calculated. \cite{khan2011automatic, ibrahim2012iris,Jan2013a} used a similar technique in which two secure regions are defined. Then their gradients are computed, followed by excluding wrong boundary points using a distance error transform. One disadvantage of these techniques is that they bias their localization of the limbic boundary in the horizontal direction. Also, these techniques may not work for an eye image having low-intensity regions. \\
\indent To resolve these issues, we adopted a scheme that first finds the true orientation of the eye in the image. First, $I_{smooth}$ is filtered using a LoG filter of $\sigma=R$. This is then thresholded to select the top 70\% of positive values. The largest connected component that overlaps the detected pupil is then selected, as shown in Fig. \ref{boundart_test}. The detected eye image is approximately ellipse shaped with major axis almost double the minor axis. The orientation of the eye is found from the orientation of the major axis of the ellipse. There are several advantages to finding the true orientation of the eye. First, if the major axis is treated as the x-axis \cite{khan2011automatic,ibrahim2012iris,Jan2013a} then the chances of getting noise in the stable zone will be very low. This will certainly increase the accuracy of these algorithms. It also makes it easier to find the  area affected by the eyelashes and eyelid (occlusion zone). Second, the orientation also facilitates the iris normalisation and matching process.\\
\indent To extract the outer boundary of the iris, the eye image is divided into four regions: left stable zone, right stable zone, upper occlusion zone, and lower occlusion zone, as shown in Fig. \ref{OuterBoundry:-a}. Usually, the upper occlusion zone affects the iris region more than lower occlusion because of the eyelashes and eyelid. The detection of these zones not only plays a vital role in limbic boundary localization but also in iris normalisation. In this chapter, this issue is addressed by first detecting the true outer boundary of the iris in the secure regions and then in the occlusion zones. To accurately detect the outer boundary of the iris, the cleaned zero-crossing image is used. Both secured zones are radially scanned in a similar manner as in \cite{khan2011automatic, Jan2013a}. Fig. \ref{secure_region} shows the iris outer boundary extracted from the secured region. The average distance of the outer boundary in the stable zones from the centre of the pupil is calculated and  is designated as the iris radius $I_{radius}$. Using this radius, the search is extended into both occlusion zones. Any discontinuity shows that the region is affected by eyelids/eyelashes, as shown in Fig. \ref{secure_region:-a}. Therefore, using these boundary points, the affected region in the iris can easily be marked, as shown in Fig. \ref{OuterBoundry:-c}. Finally, the disconnected outer circle is interpolated to give the true centre and radius of the limbic boundary, as shown in Fig. \ref{secure_region:-c}.
\section{Experimental results}
\label{ch3:sec3}
The validity of the proposed method is evaluated  on three public databases, namely: MMU version 1.0 database \cite{mmu},  CASIA-IrisV1 database \cite{CASIA} and CASIA-IrisV3-Lamp database \cite{CASIA}. The accuracy rate is used to measure the performance of the proposed method. The accuracy rate $(A_{r})$ is based on the accuracy error $A_e$, which is defined as
\begin{equation}\label{ExpEq1}
 {A_e} = \frac{{\left| {{N_a} - {N_{\det }}} \right|}}{{{N_{total}}}} \times 100,
\end{equation}
where $N_{det}$ and $N_a$  are the numbers of detected and actual iris pixels, respectively. The actual iris pixels are calculated manually as suggested in \cite{ibrahim2012iris}.  If $A_e$ is less than 10\%, then the detected iris is considered to be the true iris. $A_r$ is defined as
\begin{equation}\label{ExpEq2}
 {A_r} = \frac{{{N_{success}}}}{{{N_{total}}}} \times 100,
\end{equation}
where $N_{total}$ is the total number of images in the database and $N_{success}$ is the total number of eye images in which the iris has been successfully localized. The following sections describe the details of the experimental results.
\subsection{Experimental setup 1}
\begin{figure}[h]
 	\begin{center}
 		\subfigure[] {\label{MMU:-a}\includegraphics[scale=0.11]{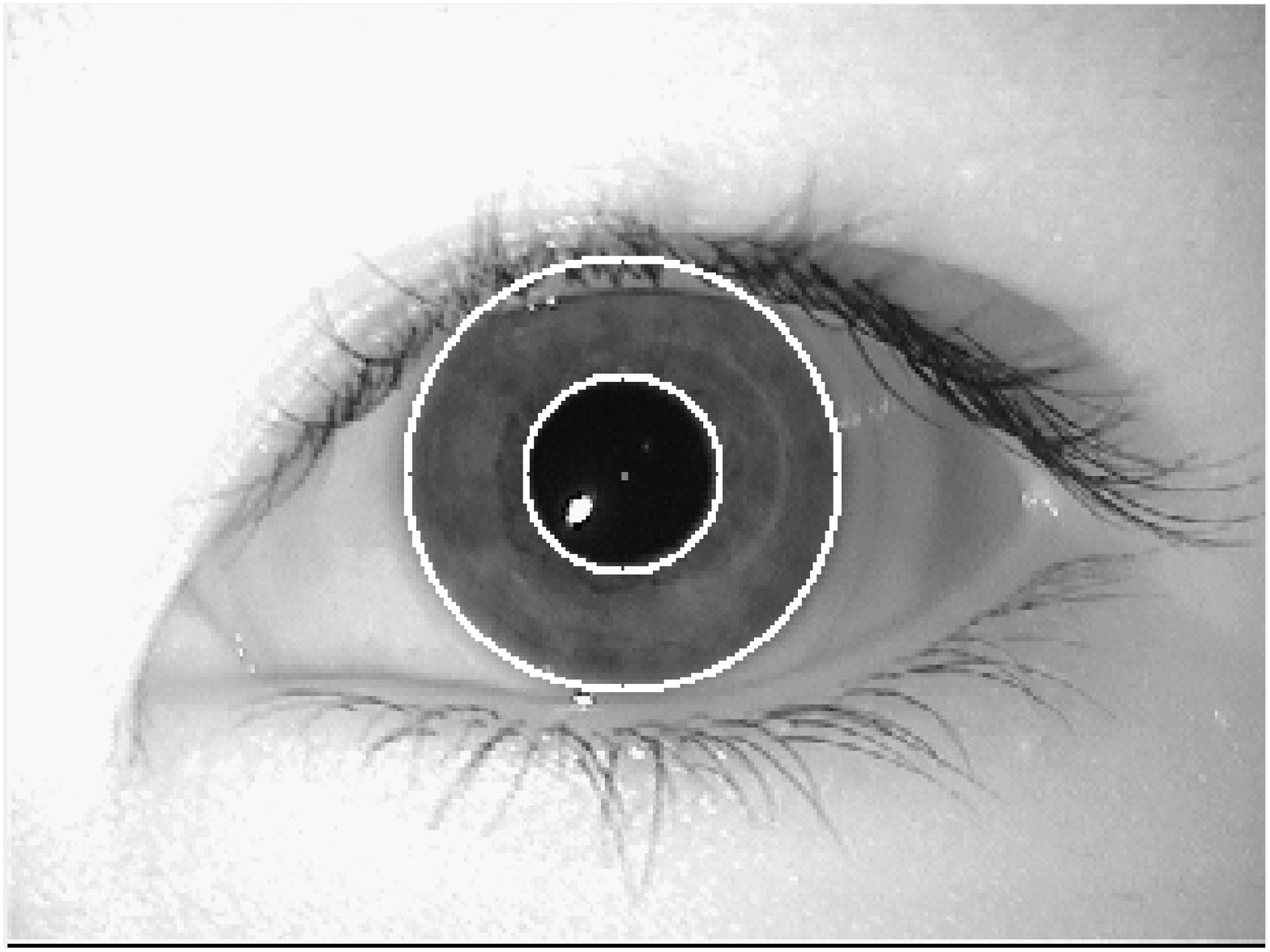}}
 		\subfigure[] {\label{MMU:-b}\includegraphics[scale=0.11]{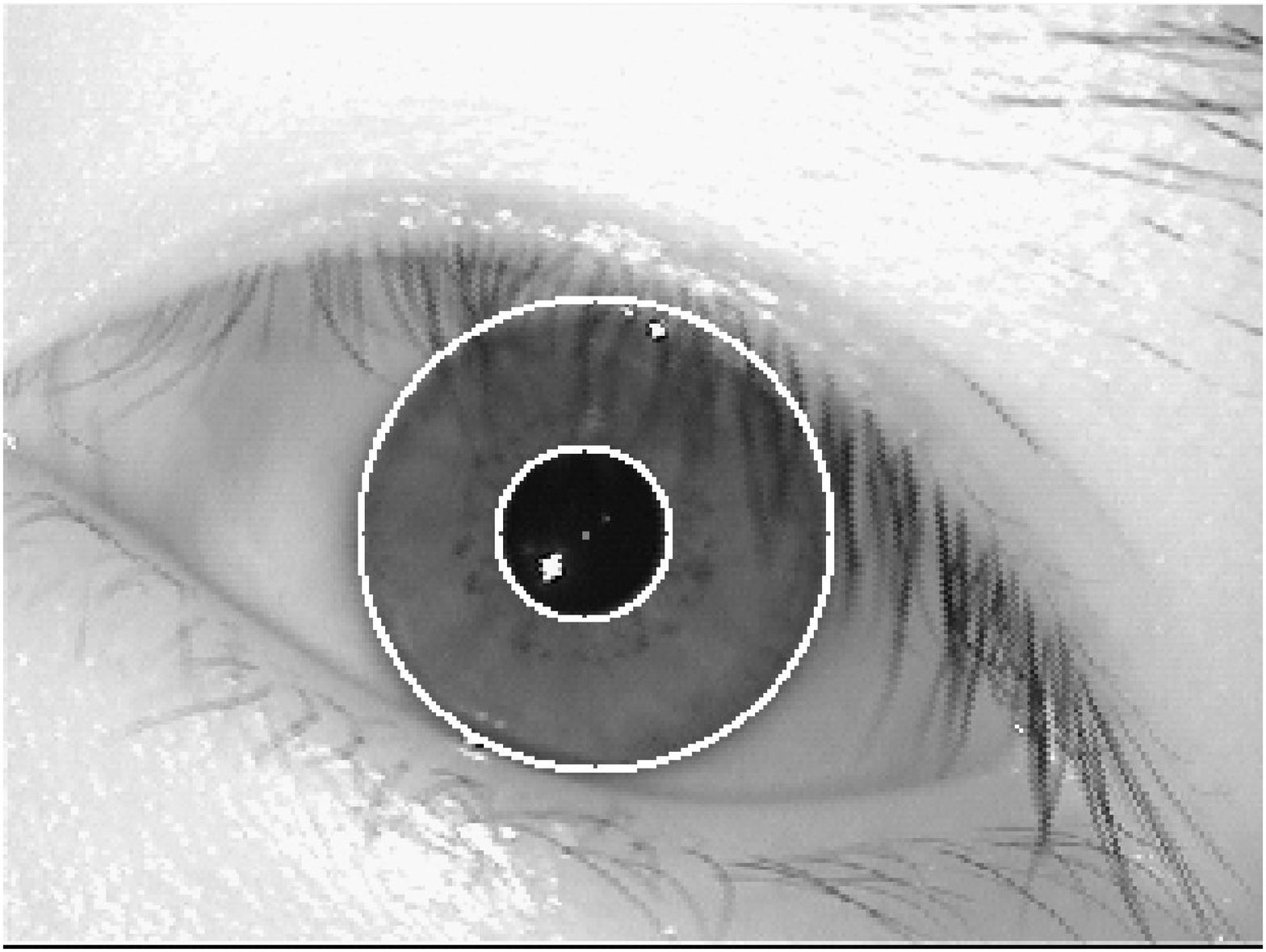}}\\
  		\subfigure[] {\label{MMU:-c}\includegraphics[scale=0.11]{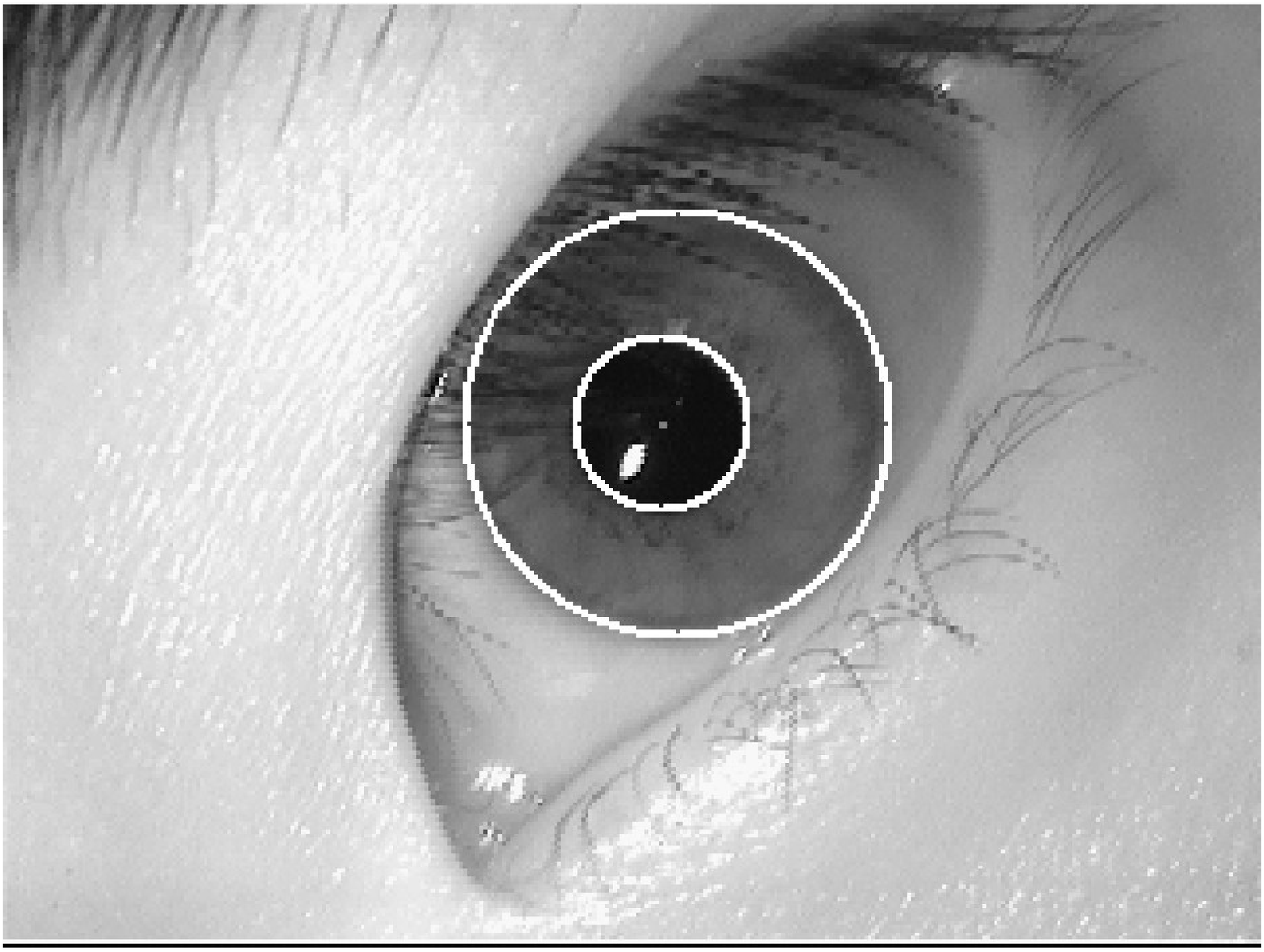}}
        \subfigure[] {\label{MMU:-d}\includegraphics[scale=0.11]{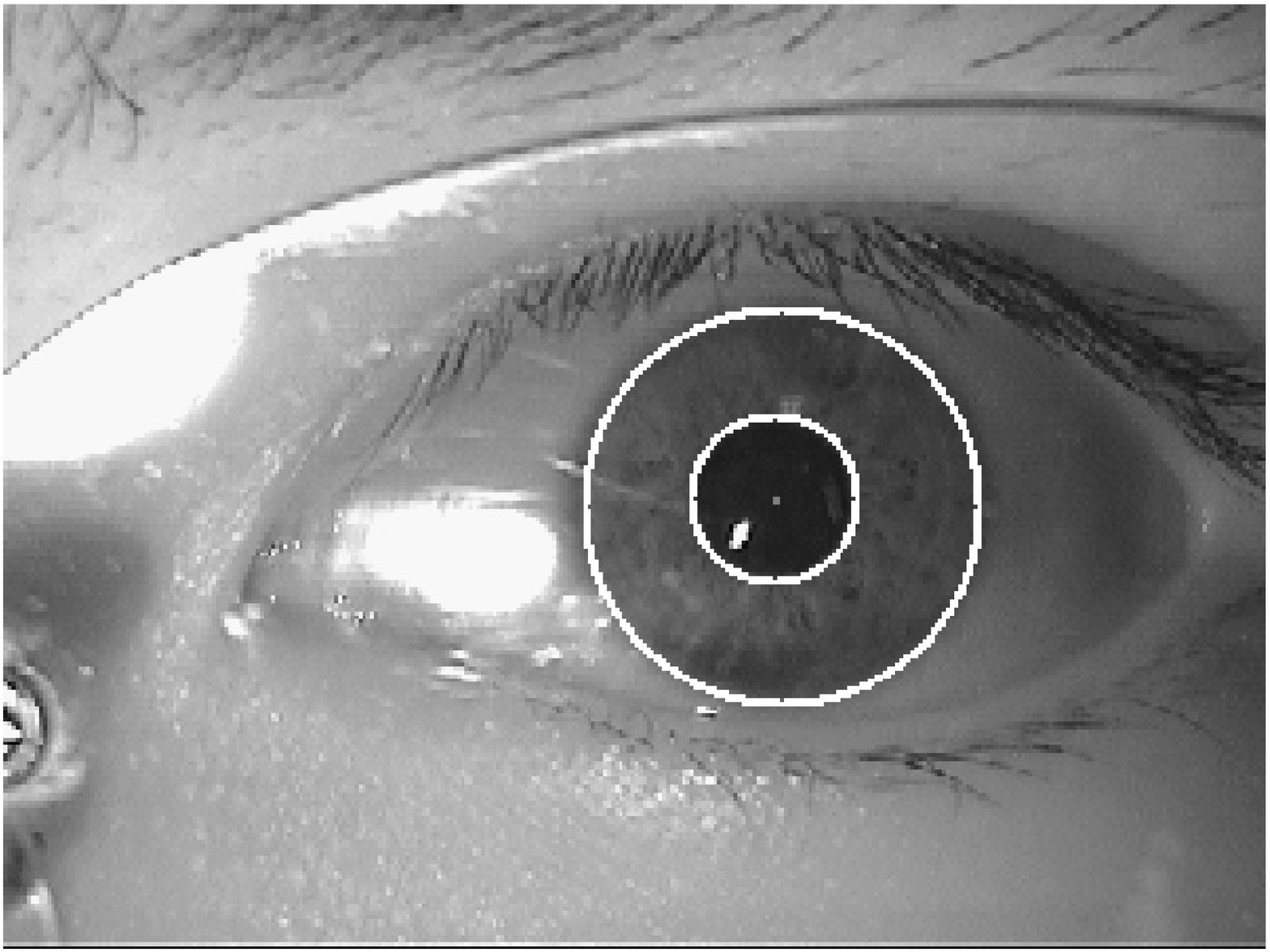}}\\
        \subfigure[] {\label{MMU:-e}\includegraphics[scale=0.11]{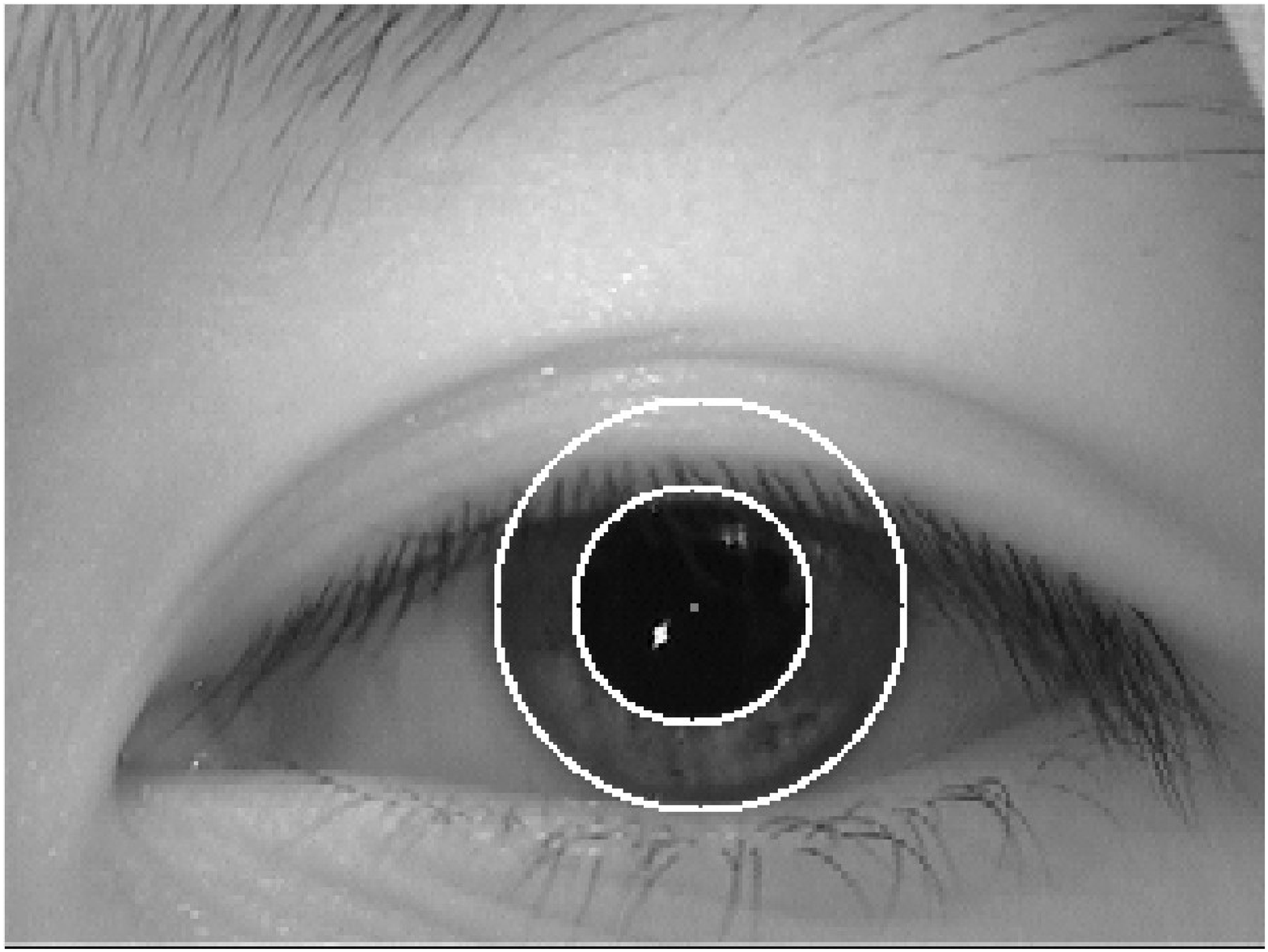}}
        \subfigure[] {\label{MMU:-f}\includegraphics[scale=0.11]{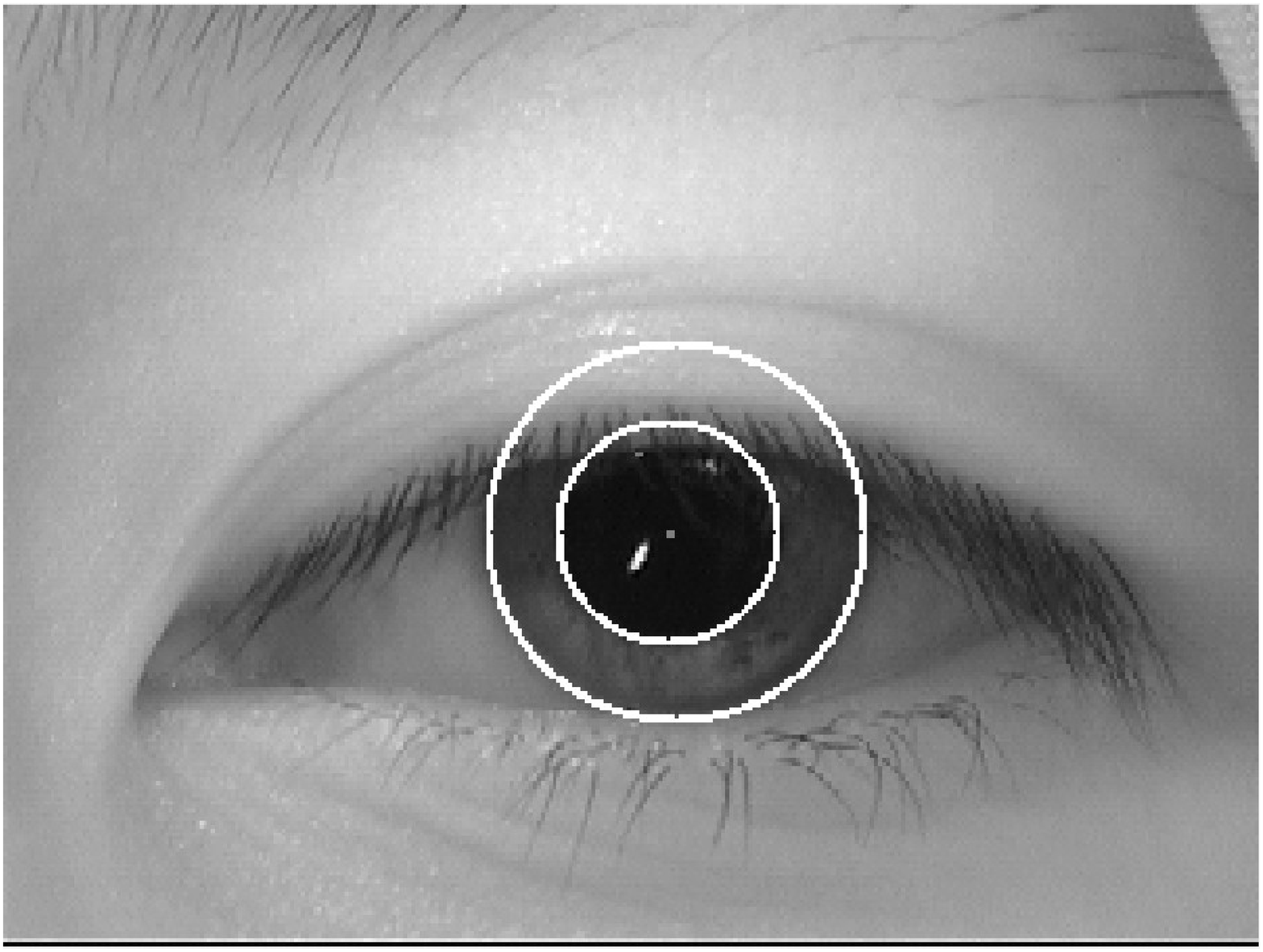}}
 	\end{center}
 	\caption{Results of proposed method on some noisy images of the MMU version 1 database.}
 	\label{MMUresults}
 \end{figure}

In this experiment, results are collected using data from the MMU version 1.0 database. This database contains 450 images of 45 subjects, i.e., 10 images per subject. The resolution of each image is 320$\times$240 pixels. The proposed method has been tested on the whole database. An accuracy rate of 100\% is achieved on both the inner and the outer boundary of the iris. Fig. \ref{MMUresults} shows the results of the proposed method on some of the randomly selected images from this. Table \ref{MMUTable} compares the accuracy of the proposed method with several existing methods on the MMU version 1.0 database.
\begin{table}
 \begin{center}
 \caption{Comparison of some recent segmentation algorithms applied to the MMU database (Results are taken from ~\cite{Dey2008}).}\label{MMUTable}
 \begin{tabular}{|p{1.7in}|p{1.7in}|} \hline
Method                                               & Accuracy \\ \hline
Masek~\cite{Masek2003}            & 83.9\% \\ \hline
Daugman~\cite{Daugm2002}    & 85.6\% \\ \hline
Ma et. al.~\cite{Ma2004}              & 91\% \\ \hline
Daugman ~\cite{Daugm2007}   & 98.2\% \\ \hline
Somanth et. al.~\cite{Dey2008} & 98.4\% \\ \hline
Proposed                                            & 100\% \\ \hline
\end{tabular}
\end{center}
\end{table}
\subsection{Experimental setup 2}
\begin{figure}[h]
 	\begin{center}
 		\subfigure[] {\label{CASIA1:-a}\includegraphics[scale=0.12]{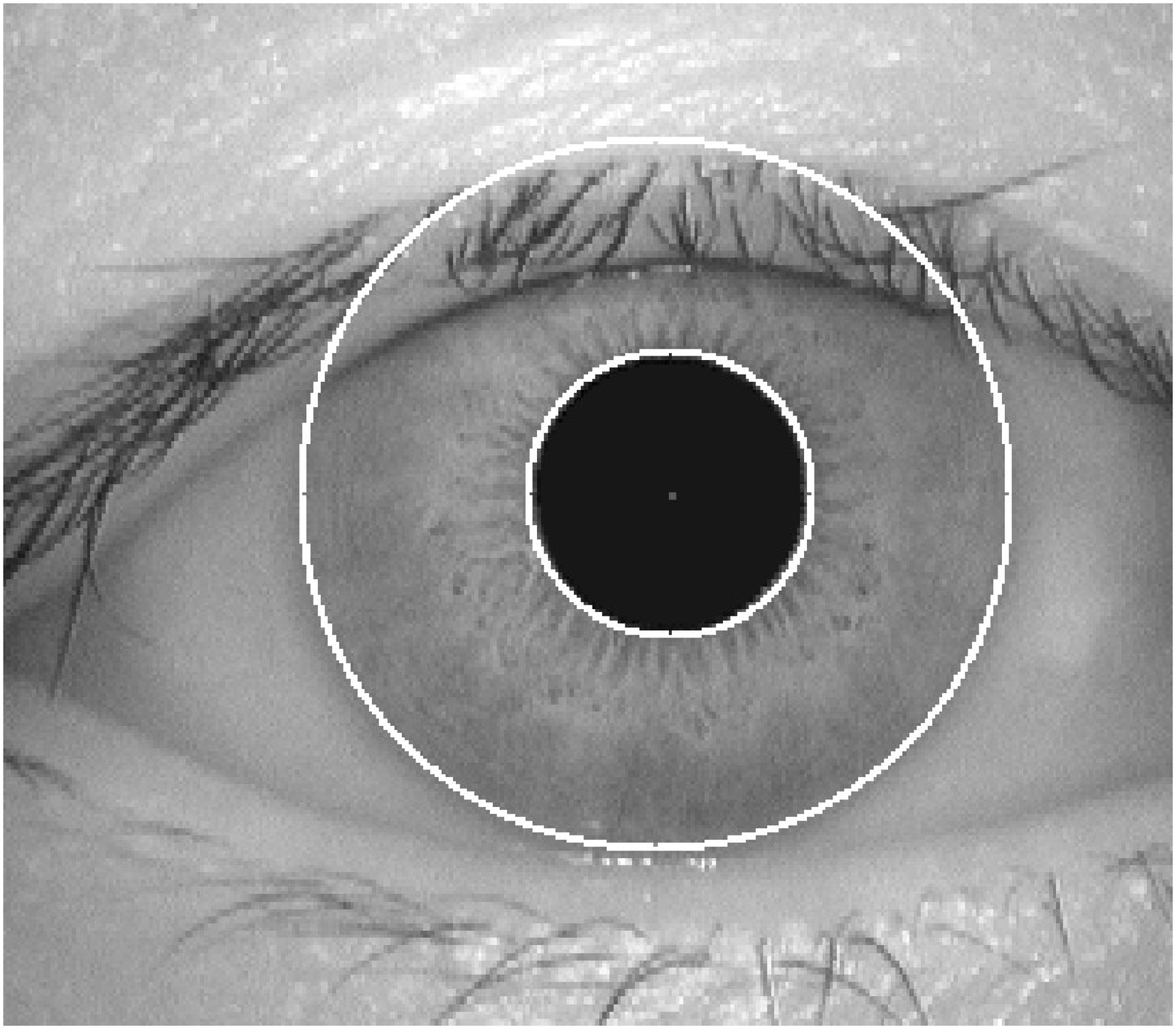}}
 		\subfigure[] {\label{CASIA1:-b}\includegraphics[scale=0.12]{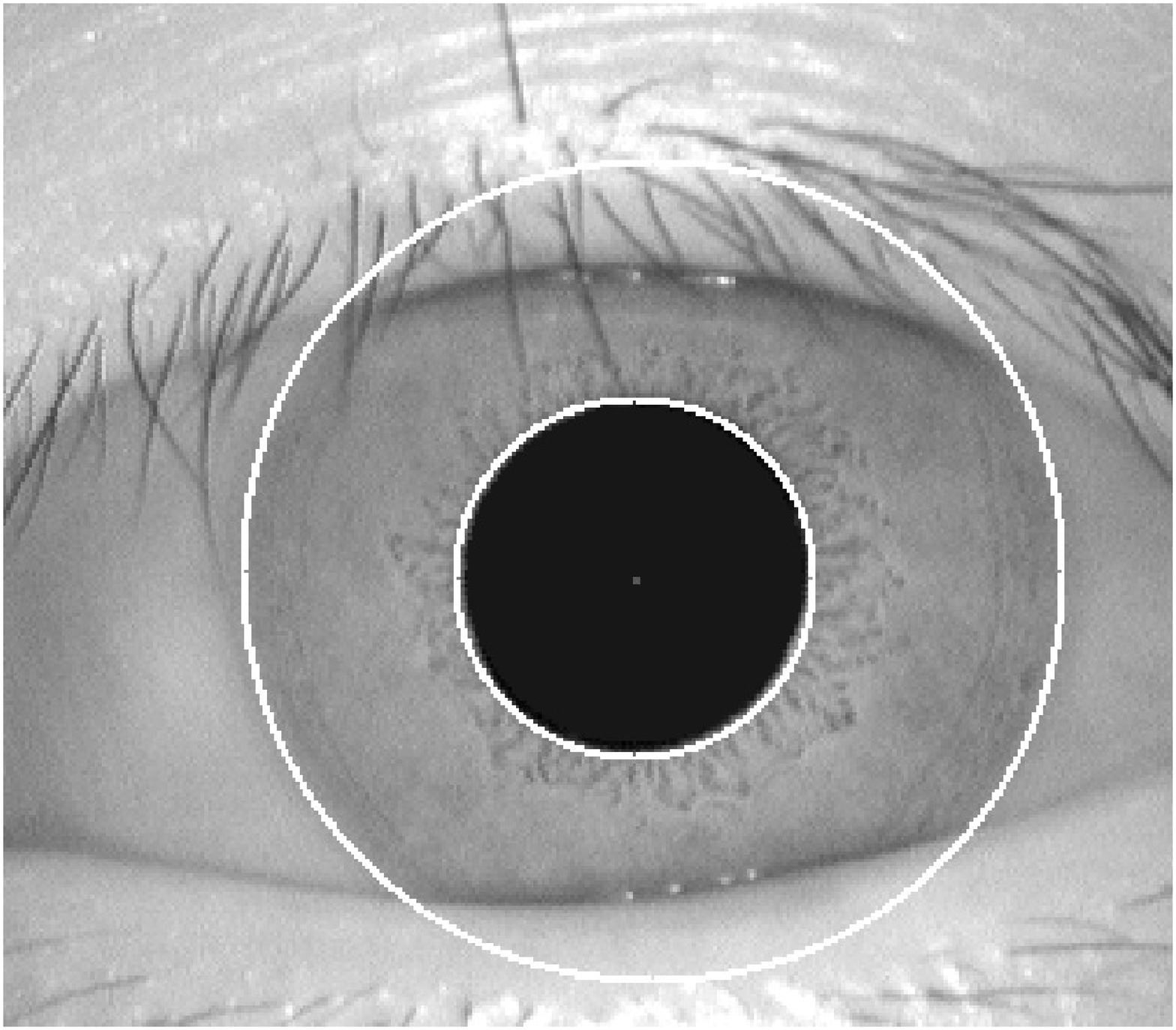}}\\
  		\subfigure[] {\label{CASIA1:-c}\includegraphics[scale=0.12]{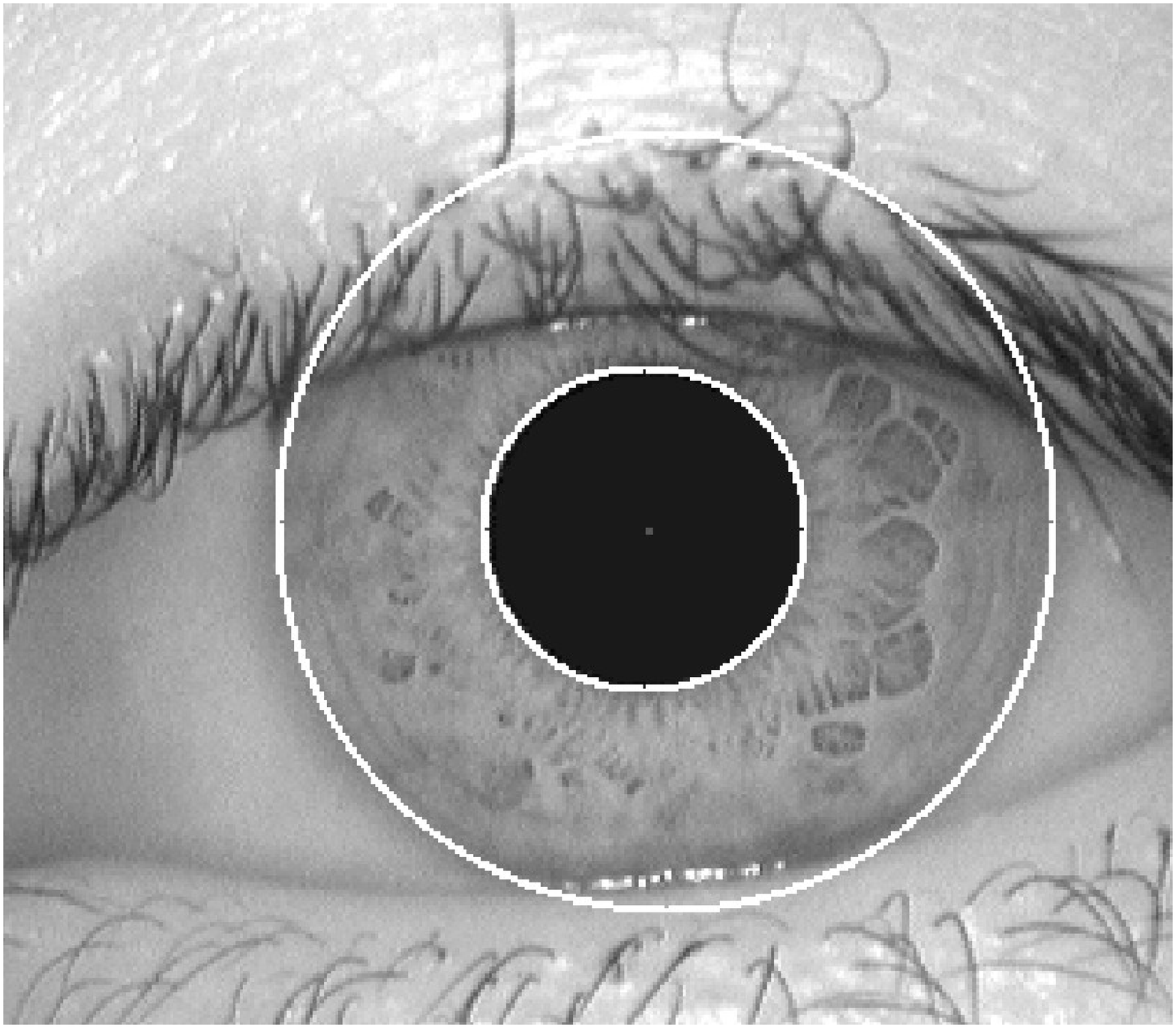}}
        \subfigure[] {\label{CASIA1:-d}\includegraphics[scale=0.12]{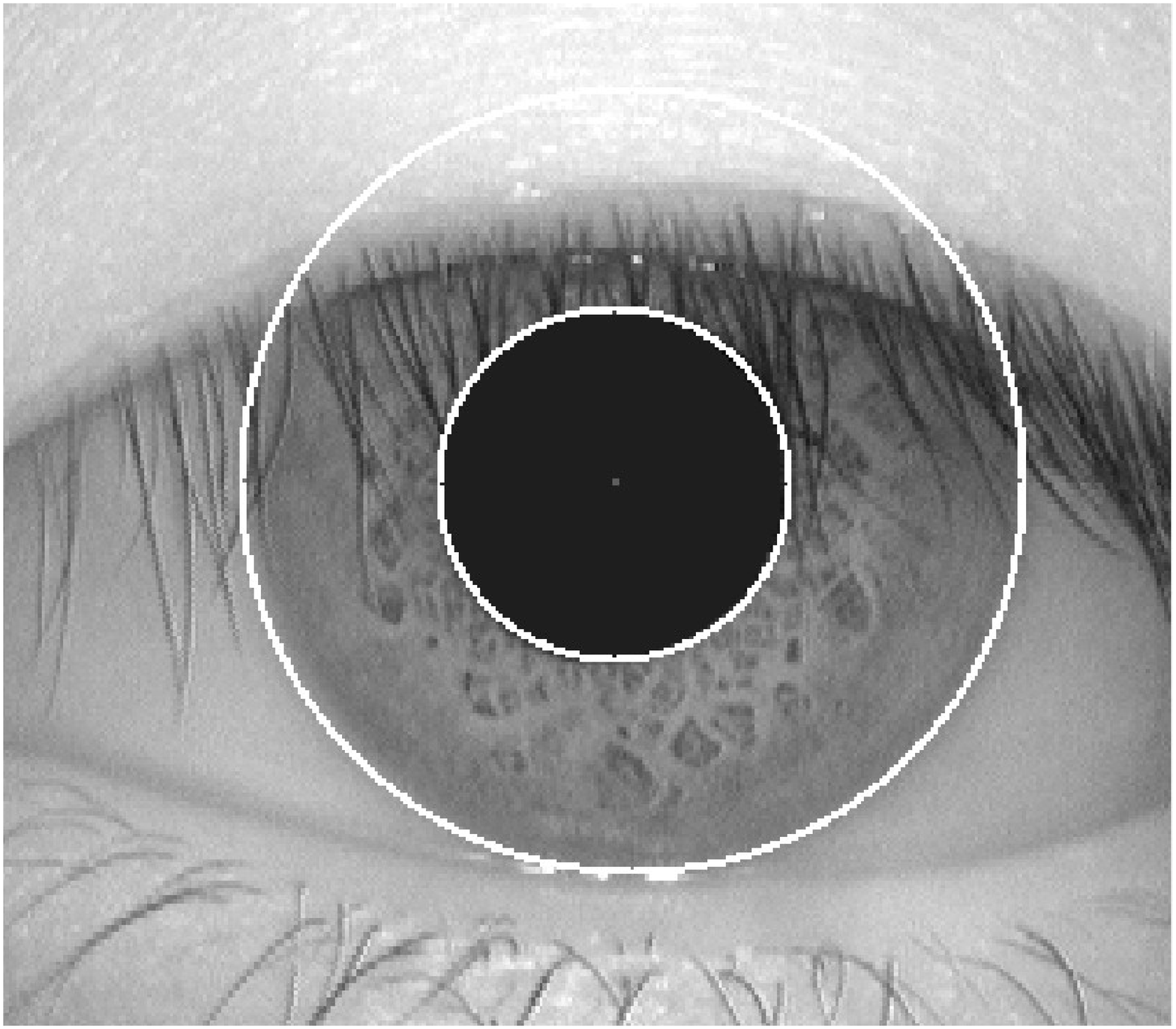}}\\
        \subfigure[] {\label{CASIA1:-e}\includegraphics[scale=0.12]{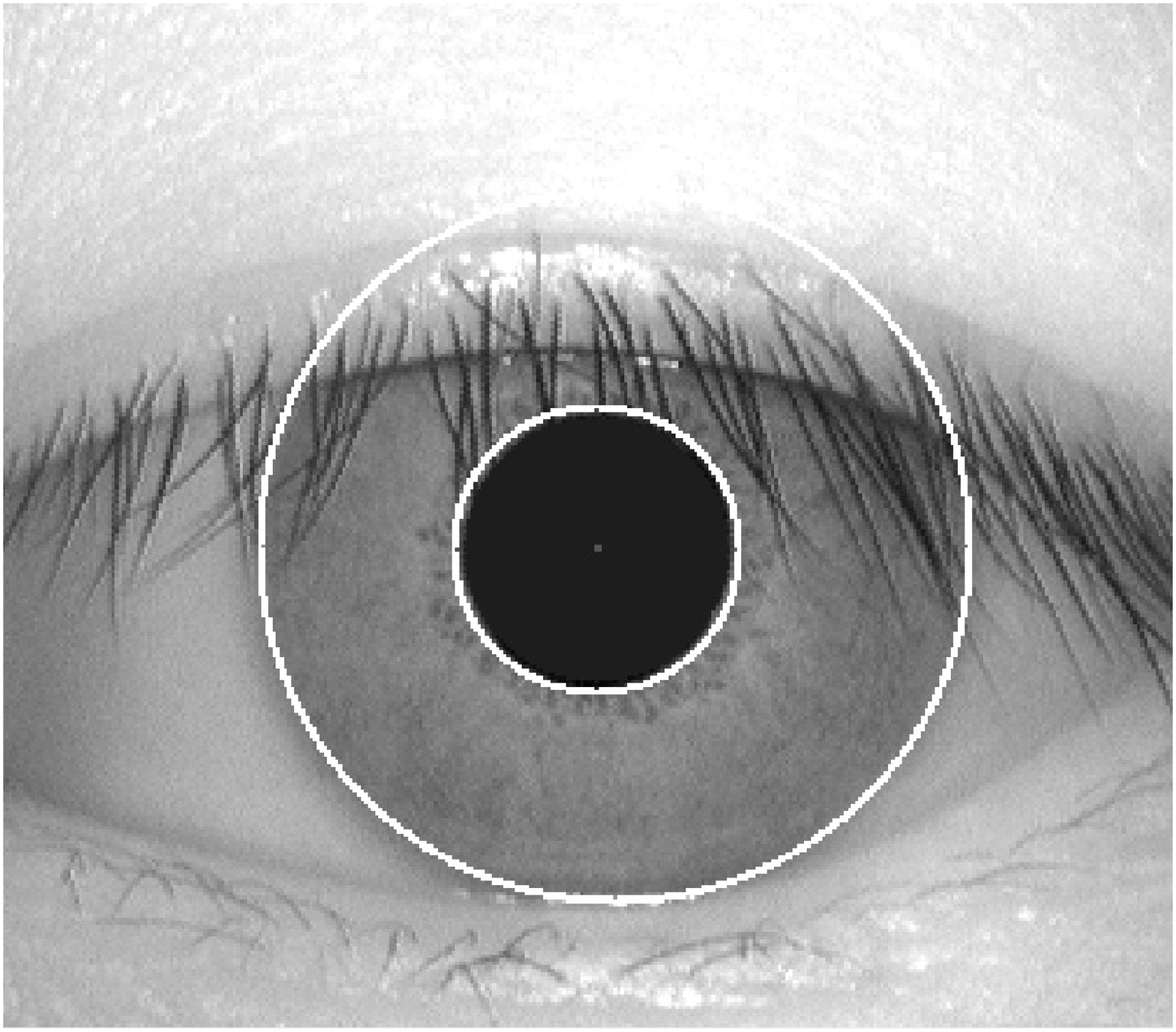}}
        \subfigure[] {\label{CASIA1:-f}\includegraphics[scale=0.12]{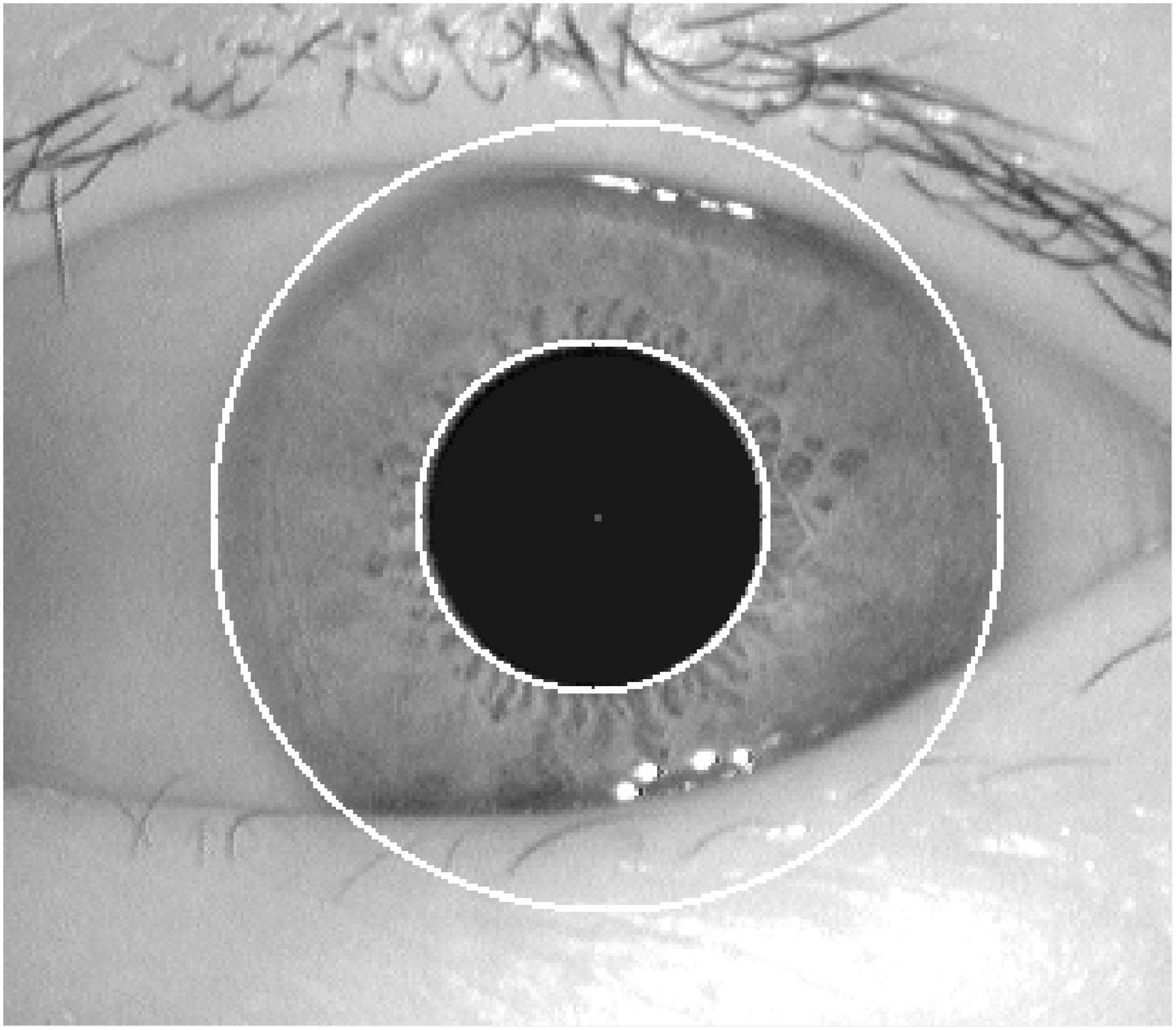}}
 	\end{center}
      	\vspace{-17pt}
 	\caption{Results of proposed method on some noisy images from the CASIA Ver 1.0 database.}
 	\label{CASIA1results}
 \end{figure}

The second experiment is performed  on the CASIA Ver 1.0 iris database. This database contains 756 eye images of 108 subjects, 7 images per individual. The resolution of each image is 320$\times$280 pixels. Using the mixture of LoG filtering and zero-crossings of LoG filter we achieve 100\% accuracy on this database for both the inner and the outer boundary of the iris. Fig. \ref{CASIA1results} shows the results of the proposed method on some of the randomly selected images from CASIA-IrisV1. Table \ref{CASIA1Table} compares the proposed method with existing methods using CASIA-IrisV1.
\begin{table}
\begin{center}
 \caption{Comparison of some recent segmentation algorithms over the CASIA 1.0 database (Results taken from the published work).}\label{CASIA1Table}
\begin{tabular}{|p{1.7in}|p{1.7in}|} \hline
Method & Accuracy \\ \hline
Mateo~\cite{Otero-Mateo2007}    & 95\% \\ \hline
Yuan,W~\cite{Yuan2005}                  & 99.45\% \\ \hline
Wildes ~\cite{Wildes1997}               & 99.9\% \\ \hline
Cui~\cite{cui_04}                                  & 99.34\% \\ \hline
Daugman~\cite{Daugm1993}        & 98.6\% \\ \hline
A.Basit~\cite{Basit2007}                   & 99.6\% \\ \hline
Proposed                                                & 100\% \\ \hline
\end{tabular}
\end{center}
\end{table}

\subsection{Experimental setup 3}
\begin{figure}[h]
 	\begin{center}
 		\subfigure[] {\label{CASIA3:-a}\includegraphics[scale=0.15]{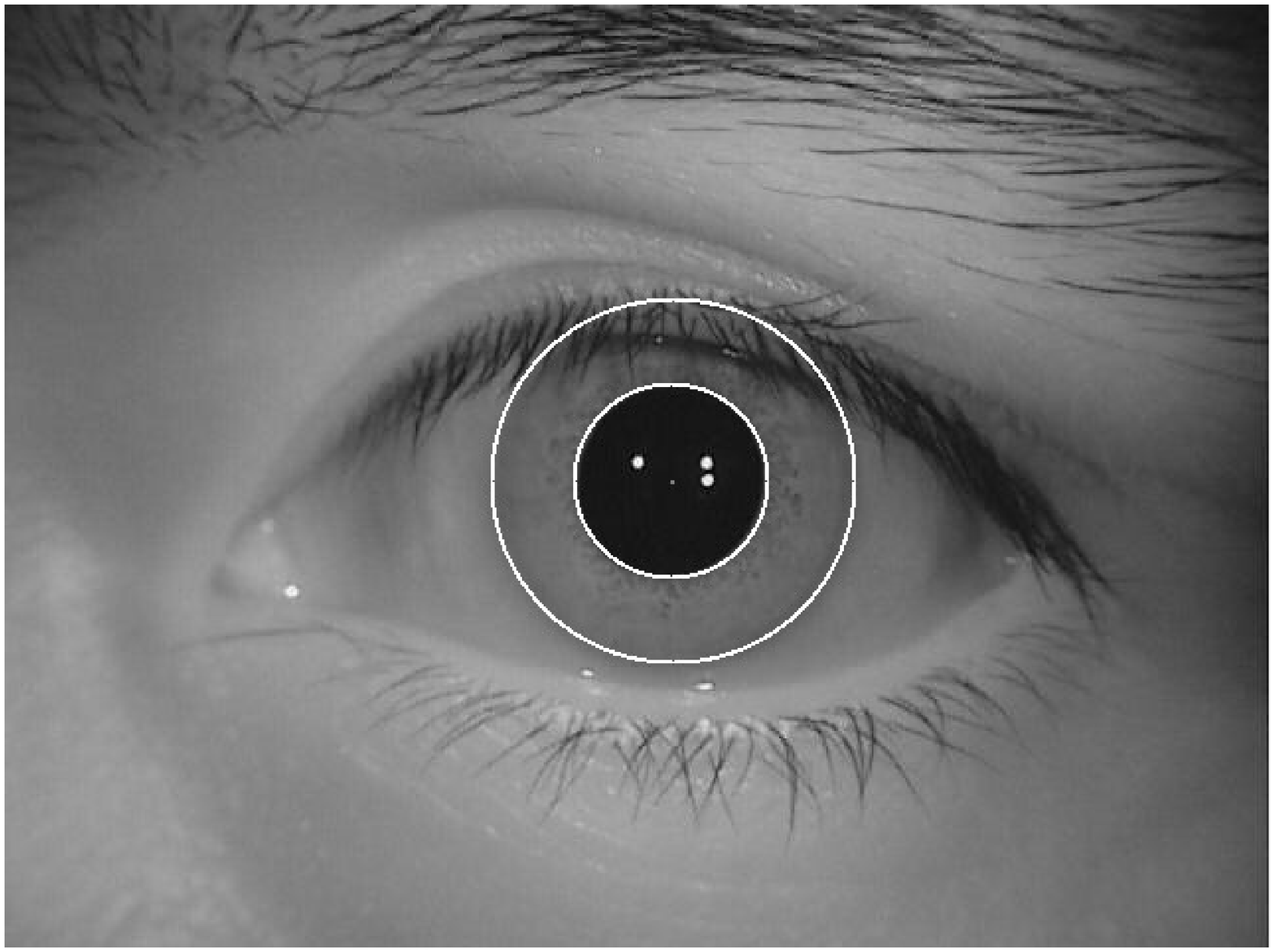}}
 		\subfigure[] {\label{CASIA3:-b}\includegraphics[scale=0.15]{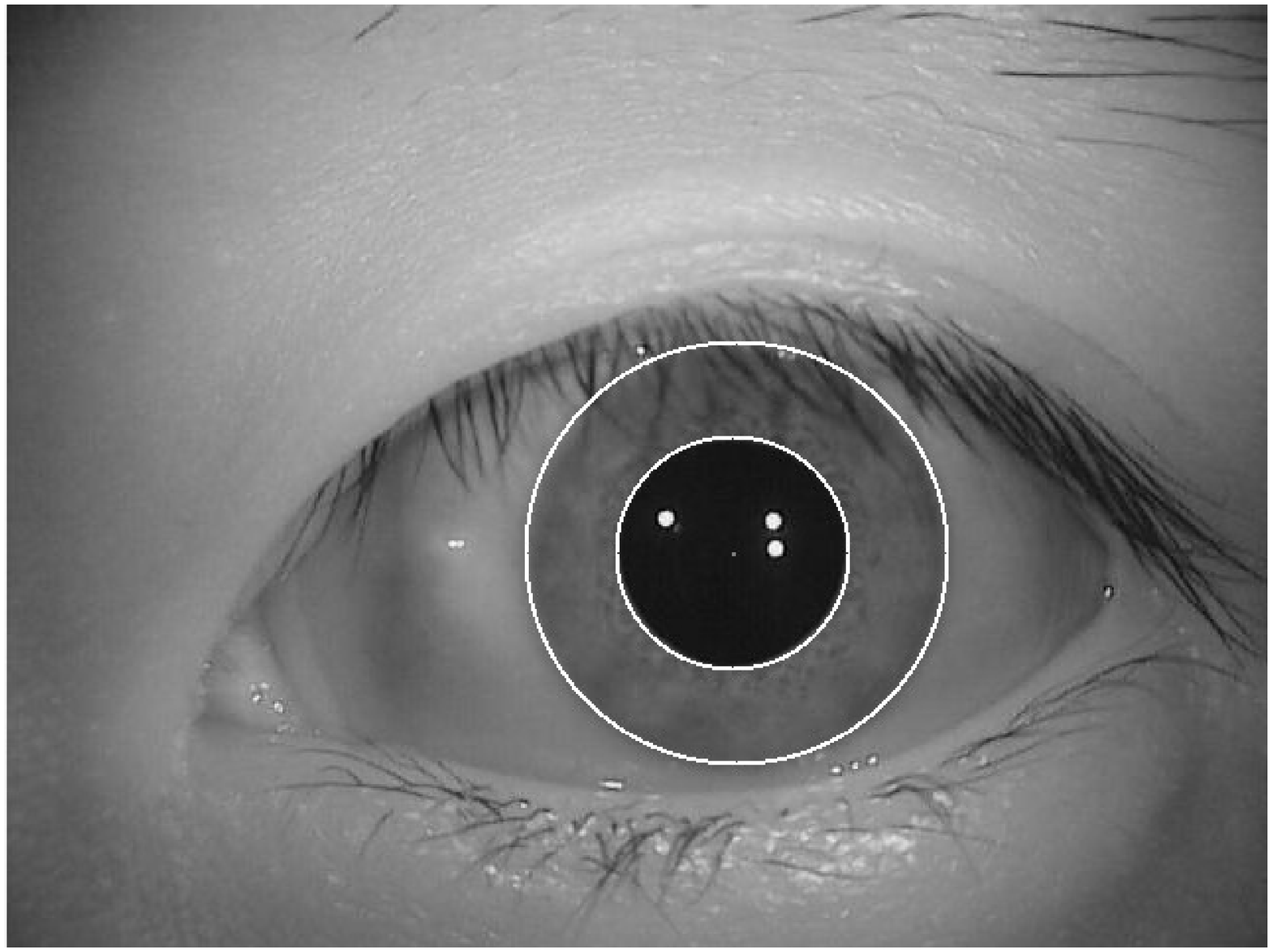}}\\
  		\subfigure[] {\label{CASIA3:-c}\includegraphics[scale=0.15]{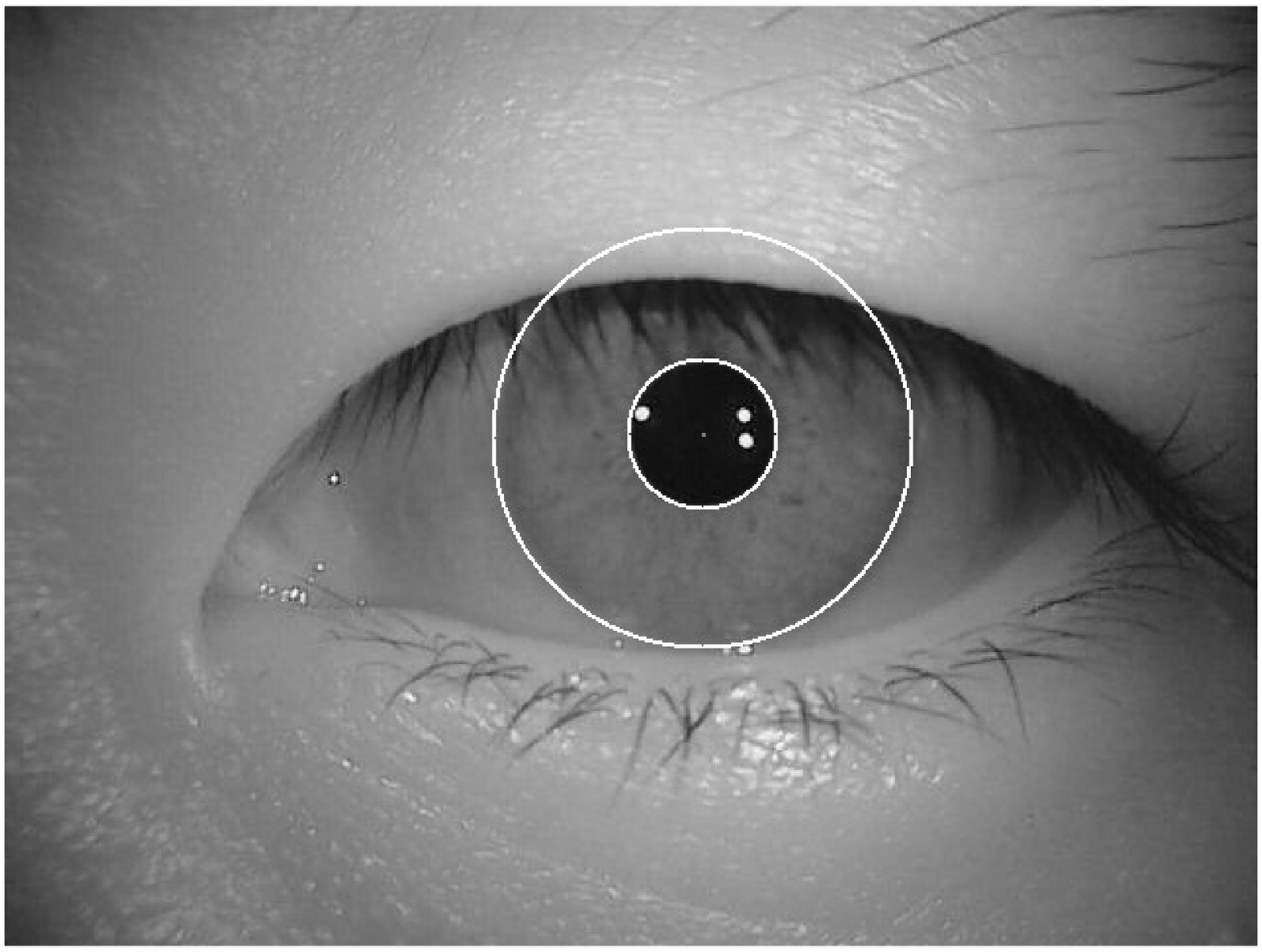}}
        \subfigure[] {\label{CASIA3:-d}\includegraphics[scale=0.15]{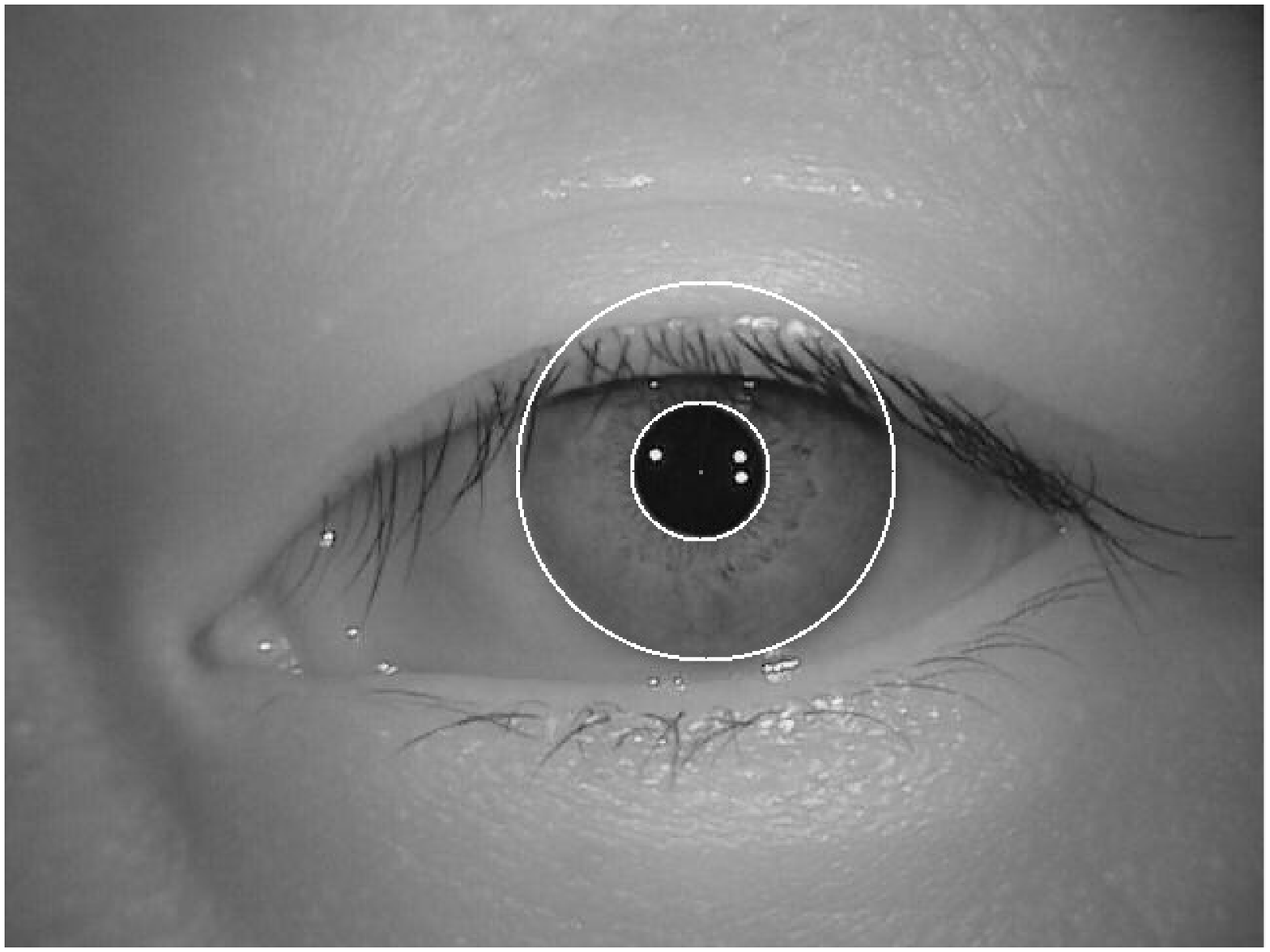}}\\
        \subfigure[] {\label{CASIA3:-c}\includegraphics[scale=0.15]{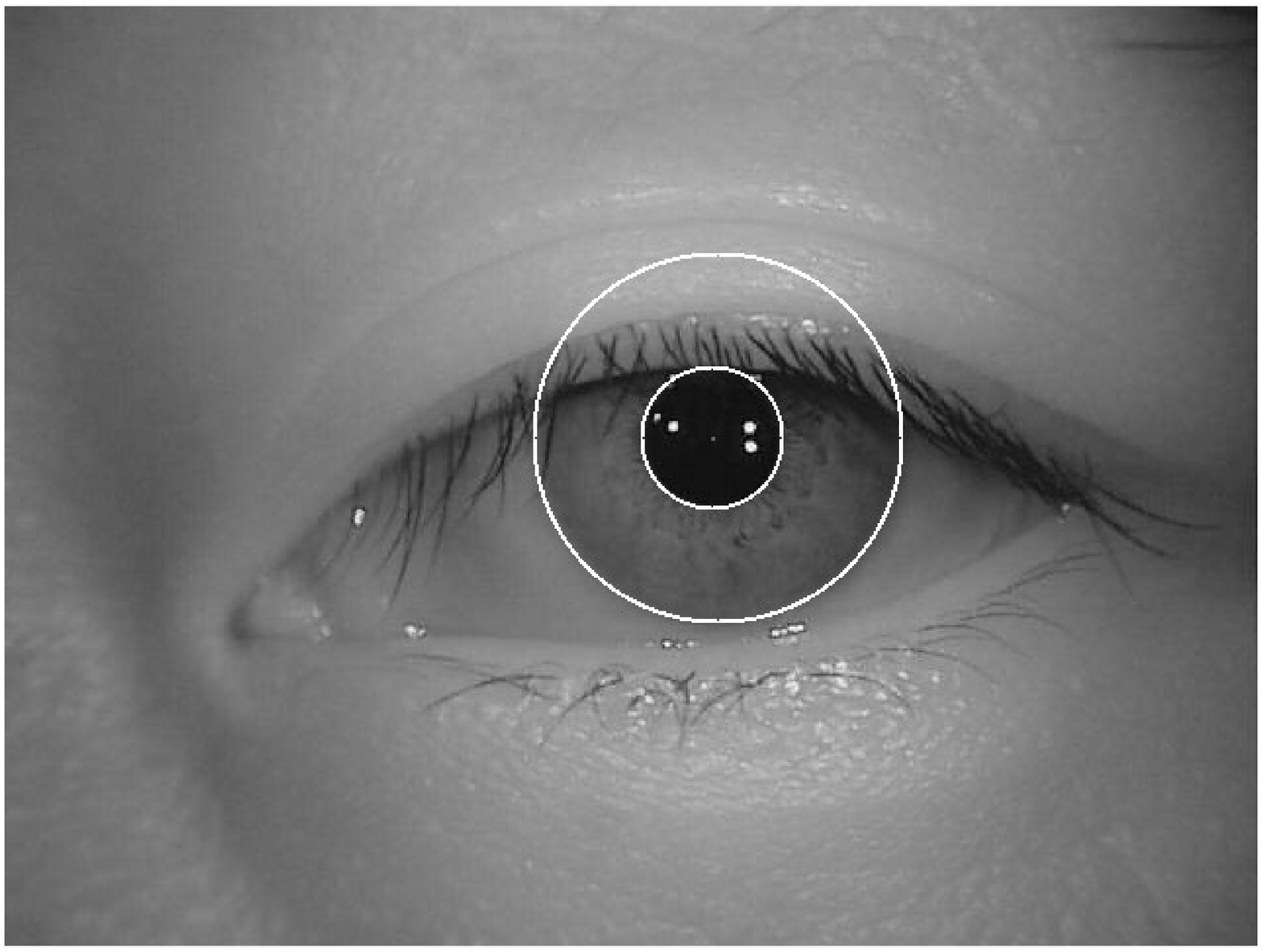}}
        \subfigure[] {\label{CASIA3:-d}\includegraphics[scale=0.15]{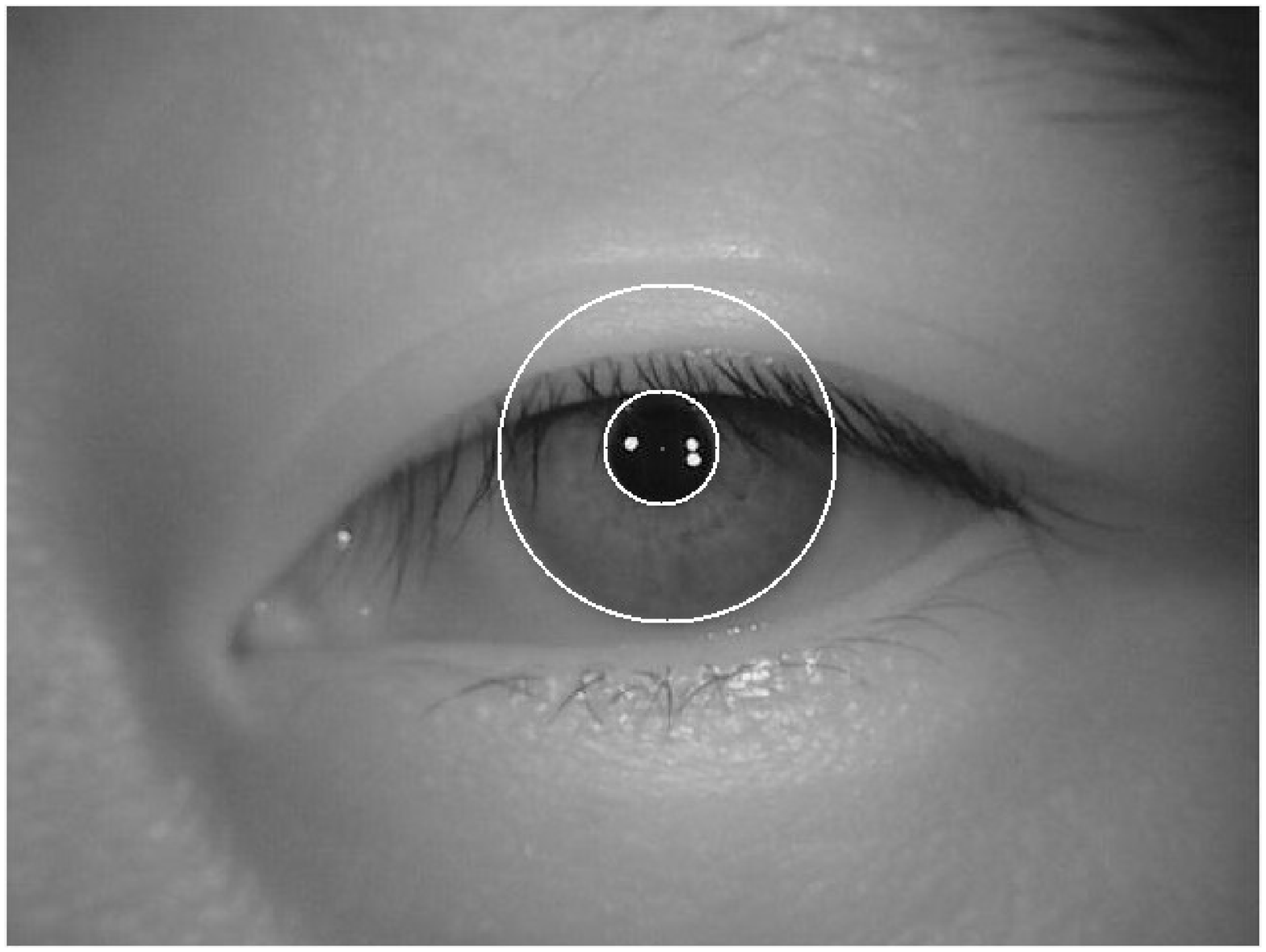}}
 	\end{center}
      	\vspace{-17pt}
 	\caption{Results of proposed method on some noisy images from the CASIA-IrisV3-Lamp database.}
 	\label{CASIA3results}
 \end{figure}

In this experiment, results are collected using data from the CASIA-IrisV3-Lamp iris database. This database contains 16440 images of 441 subjects. Each subject contributed 40 images, 20 images of the each eye with a resolution of 640$\times$480 pixels. The proposed method was tested on images from the first 102 subjects. The accuracy rate using the CASIA-IrisV3-Lamp database is 99.68\%. Table \ref{CASIA3Table} shows the accuracy rate of the proposed method on CASIA-IrisV3-Lamp database. Fig. \ref{CASIA3results} shows the results on some randomly selected images from the CASIA-IrisV3-Lamp database.
\begin{table}
\begin{center}
 \caption{Comparison of some recent segmentation algorithms using the CASIA-IrisV3-Lamp (Results taken from~\cite{ibrahim2012iris}).}
   \label{CASIA3Table}
\begin{tabular}{|p{1.7in}|p{1.7in}|} \hline
Method                                                        & Accuracy \\ \hline
Masek~\cite{Masek2003}                     & 79.02\% \\ \hline
Ibrahim~\cite{ibrahim2012iris}           & 98.28\% \\ \hline
Proposed                                                    & 99.55\% \\ \hline
\end{tabular}
\end{center}
\end{table}

 \section{Computational cost and limitations}
 \label{ch3:sec4}
\begin{figure}
  \centering
  \includegraphics[scale=0.4]{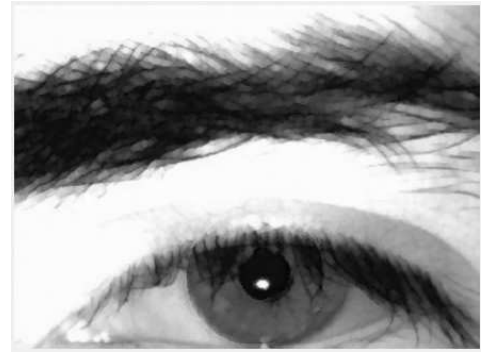}\\
  \caption{Iris image severally occluded by the eyelashes and eyebrows.}\label{Fail}
\end{figure}

 The average computational cost is computed for 100 randomly selected eye images from the MMU version 1.0, CASIA Ver 1.0, and CASIA-IrisV3-Lamp databases.  MATLAB built-in facility is utilised to obtain the optimal results. Table \ref{Computation} shows a comparison on the computational cost of the proposed method and three similar states of the art method. \\
\indent  In the presence of dense eyelashes and eyebrows, as shown in Fig. \ref{Fail}, the proposed method fails to locate the true pupillary boundary. Such dense eyebrows and eyelashes effect the LoG filtering along with the region growing. Although, this can be handled by setting a different $\sigma$ for LoG filter and tuning the region growing, for the generic parameters that are set for the whole database, it fails to locate true pupillary boundary. Similarly, if the iris region is affected with dense eyelashes then a single value of ${\lambda _c}$ fail to suppress the effect of eyelashes in the iris region.

 \begin{table*}
\begin{center}
  \caption{Processing speed (in seconds) comparison of the proposed with some of the existing methods}
    \label{Computation}
\begin{tabular}{|p{1.2in}|p{1.5in}|p{1.1in}|p{1.6in}|}\hline
Method                                             & MMU version 1.0 & CASIA Ver 1& CASIA-IrisV3-Lamp \\ \hline
Ibrahim~\cite{ibrahim2012iris} & 0.95                          &1.1      & 2.2 \\ \hline
Jan~\cite{Jan2013}        & 1.5                            & 1.7      &3.0 \\ \hline
Labati~\cite{Labati2010}            & 2.6                            &3.0       & Not available \\ \hline
Proposed                                         & 0.44                          & 0.5      &1.4 \\ \hline
\end{tabular}
\end{center}

\end{table*}
\section{Conclusion}
\label{ch3:sec5}
This chapter presents a fast and an accurate iris segmentation technique for iris biometrics. There are four major contributions. First, we develop a fast and novel method for pupil segmentation that is based on a shape detector and an intensity-based threshold. The use of a LoG filter followed by region growing gives an estimate of the pupil centre and radius. \\
\indent The true pupillary boundary is refined using the zero-crossings of a second LoG filter. Next, the true orientation of the eye in the image is estimated using a third LoG filter. The orientation facilitates the process of locating the true limbic boundary of the iris and eyelids. Using the zero-crossings of the LoG, the search is initially started from the stable zones and then extended to the  occlusion zones. The discontinuities are located, which give indications of eyelids in the iris region. Finally, using the interpolation the iris outer boundary as well as eyelid arcs are estimated. The proposed method also works well in estimating the inner and outer boundary of the iris in the case of partially opened eye images and scattered eyelashes.\\
\indent The extensive experimental results on three iris databases show that the proposed method is computationally less expensive in achieving the  state-of-the-art iris segmentation accuracy.

\section*{Acknowledgments}
The authors wish to thank the Chinese Academy of Science-Institute of Automation for providing CASIA ver 1.0 and CASIA ver 3.0 iris database. The author also wish to thank the Multimedia University for providing MMU iris database.
\bibliographystyle{IEEEtran}
\bibliography{tariq_citations}
\end{document}